\documentclass[10pt]{article} 

\usepackage[preprint]{rlj}           
\usepackage{algorithm}
\usepackage{algorithmic}
\usepackage{subcaption}
\usepackage{booktabs}
\usepackage{wrapfig}

\usepackage{amssymb}            
\usepackage{mathtools}          
\usepackage{mathrsfs}           
\usepackage{graphicx}           
\usepackage{subcaption}         
\usepackage[space]{grffile}     
\usepackage{url}                
\usepackage{lipsum}             
\usepackage{mathrsfs} 

\title{Causal-Paced Deep Reinforcement Learning}

\setrunningtitle{Causal-Paced Deep Reinforcement Learning}

\author{
  Geonwoo Cho \quad
  Jaegyun Im \quad
  Doyoon Kim \quad
  Sundong Kim
}

\emails{\{gwcho.public, jaegyun.public, dykim.research\}@gmail.com, sundong@gist.ac.kr}

\affiliations{
\textbf{Gwangju Intsitute of Science and Technology}\\
\par 
}

\begin{document}

\maketitle  

\begin{abstract}
Designing effective task sequences is crucial for curriculum reinforcement learning (CRL), where agents must gradually acquire skills by training on intermediate tasks. A key challenge in CRL is to identify tasks that promote exploration, yet are similar enough to support effective transfer. While recent approach suggests comparing tasks via their Structural Causal Models (SCMs), the method requires access to ground-truth causal structures, an unrealistic assumption in most RL settings. In this work, we propose Causal-Paced Deep Reinforcement Learning (CP-DRL), a curriculum learning framework aware of SCM differences between tasks based on interaction data approximation. This signal captures task novelty, which we combine with the agent’s learnability, measured by reward gain, to form a unified objective. Empirically, CP-DRL outperforms existing curriculum methods on the Point Mass benchmark, achieving faster convergence and higher returns. CP-DRL demonstrates reduced variance with comparable final returns in the Bipedal Walker-Trivial setting, and achieves the highest average performance in the Infeasible variant. These results indicate that leveraging causal relationships between tasks can improve the structure-awareness and sample efficiency of curriculum reinforcement learning. We provide the full implementation of CP-DRL to facilitate the reproduction of our main results at \url{https://github.com/Cho-Geonwoo/CP-DRL}.
\end{abstract}

\section{Introduction}
\label{sec:introduction}

Just as a child first learns to crawl before walking and running, intelligent behavior in complex environments is rarely acquired in a single leap. Instead, learning unfolds through a gradual accumulation of simpler skills that scaffold more advanced capabilities. This principle underlies the idea of curriculum learning in reinforcement learning (RL), where agents are trained on a structured sequence of tasks that progressively increase in complexity~\citep{narvekar2020curriculum,florensa2018automatic, klink2022currot}. By mastering simpler subtasks before facing the target task, the agent can avoid inefficient exploration and accelerate learning.

A core challenge in curriculum reinforcement learning (CRL) is to measure how tasks differ, identify what the agent has not yet learned, and expose it to novel yet transferable tasks~\citep{hughes2024open}. Existing approaches typically approximate task differences using agent-centric signals, such as regret~\citep{jiang2021prioritized} or value disagreement~\citep{zhang2020automatic}, but these are inherently policy-dependent and sensitive to noise. As a more principled alternative,~\citet{li2024causallyaligned} propose comparing tasks via differences in their Structural Causal Models (SCMs), enabling a policy-independent and structure-aware measure of transferability.

\begin{figure}[h]
    \centering
    \includegraphics[width=0.8\linewidth]{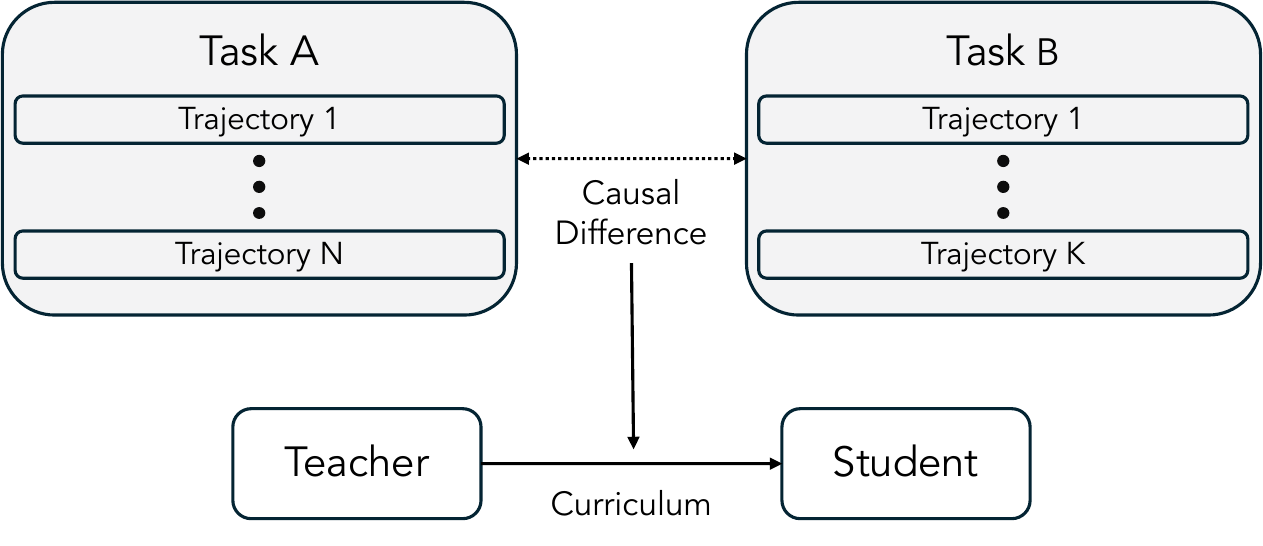}
    \caption{
    \textbf{An overview of the CP-DRL.} We estimate the causal difference between tasks based on the observed trajectories in each task. This structural signal is used by the teacher to construct a curriculum that gradually exposes the student to novel tasks. 
    }
    \label{fig:idea}
    \vspace{-1em}
\end{figure}

While SCM-based comparison provides a principled way to quantify task differences, it relies on access to the true causal structure, which is rarely available in realistic RL environments~\citep{zanga2022survey}. In this work, we propose \textbf{Causal-Paced Deep Reinforcement Learning (CP-DRL)}, a framework that addresses this limitation by approximating the SCM difference between tasks using only interaction data. Specifically, we estimate the differences in state, action, transition, and reward components, the key constituents of SCM difference in deterministic reinforcement learning, by measuring the disagreement across modular ensemble models. These component-wise disagreements are then aggregated into a causal misalignment score, which serves to identify tasks whose underlying causal structures remain unfamiliar to the agent. To balance novelty with learnability, we combine the causal misalignment score with the agent’s reward improvement, yielding a unified signal that characterizes both structural novelty and learning potential. This unified signal is then incorporated into an optimal transport-based curriculum optimization framework~\citep{klink2022currot}, enabling the curriculum to prioritize tasks that are structurally informative yet learnable, and guide training smoothly toward the target distribution. This process is conceptually illustrated in Figure~\ref{fig:idea}.

We conduct experiments on two RL benchmarks, Point Mass (PM) and Bipedal Walker (BW), demonstrating that CP-DRL effectively balances exploration and transfer through causal reasoning. In the PM environment, CP-DRL achieves the highest return and converges faster than baseline methods. In the Bipedal Walker-Trivial setting, it achieves comparable final performance with reduced variance, and in the Infeasible setting, it achieves the highest mean return among all methods. These results highlight CP-DRL’s potential to generate structure-aware curricula that enhance generalization and sample efficiency in complex reinforcement learning environments.

\section{Preliminary}
\label{sec:preliminary}

\subsection{Markov Decision Process and Contextual RL for curriculum learning}
We assume each task or environment can be modeled as a Markov Decision Process (MDP) $\mathcal{M}$, defined by the tuple $\langle \mathcal{S}, \mathcal{A}, p, r, \gamma \rangle$, where $\mathcal{S}$ is the state space, $\mathcal{A}$ the action space, $p(s' \mid s, a)$ with $s,s' \in \mathcal{S}$ and $a \in \mathcal{A}$ the transition probability function, $r(s,a)$ the reward function, and $\gamma \in [0,1)$ the discount factor. The solution to an MDP is an optimal policy $\pi(a|s)$ that maximizes the expected return $J(\pi) = \mathbb{E}_{\pi}\left[\sum_{t=0}^{\infty}\gamma^t 
r(s_t, a_t)\right]$. In some curriculum learning scenarios, tasks are parameterized or varied by some additional context $c\in \mathcal{C}$, where $\mathcal{C}$ denotes the space of possible contexts~\citep{hallak2015contextual}. For example, $c$ could represent the goal location in a navigation task, the layout of a maze, or physical properties of the environment. We can formalize a family of tasks as a Contextual MDP $\mathcal{M}(c) = \langle \mathcal{S}, \mathcal{A}, p_c, r_c, \gamma \rangle$
 where $c$  influences the transition dynamics $p_c$ and reward function $r_c$. Each value of $c$ corresponds to a different MDP, but typically there is structure shared across contexts (e.g., common state and action spaces, and some common dynamics). 
 
 The curriculum learning problem can then be formulated as,

\vspace{-2em}

\begin{equation}
\langle q_0^*, q_1^*, \dots, q_T^* \rangle = \underset{\langle q_0, q_1, \dots, q_T \rangle}{\arg\max} \; J\bigl(\pi_T\bigr)
\quad\text{where}\quad
\pi_{t+1} = \mathrm{Update}(\pi_t, q_t)
\quad\text{and}\quad q_T=\mu.
\end{equation}

where $T$ is a total number of curriculum steps, $q_t$ is the context distribution used at training step $t$, $\pi_t$ is the agent’s policy after $t$ updates, $\mu$ is a target task distribution and $\mathrm{Update}(\pi_t, q_t)$ denotes the learning step to get next policy $\pi_{t+1}$ using current policy $\pi_{t}$ and the data sampled from $q_t$. The final policy $\pi_T$ is trained via the sequence $(q_0, \dots, q_T)$ and is evaluated on $\mu$. The goal is to find a curriculum $(q_0, \dots, q_T)$ that yields a policy $\pi_T$ performing well on the target distribution.

\subsection{Structural Causal Model in RL}

\begin{wrapfigure}{r}{0.45\textwidth} 
    \vspace{-3.5em}
    \centering
    \includegraphics[width=0.7\linewidth]{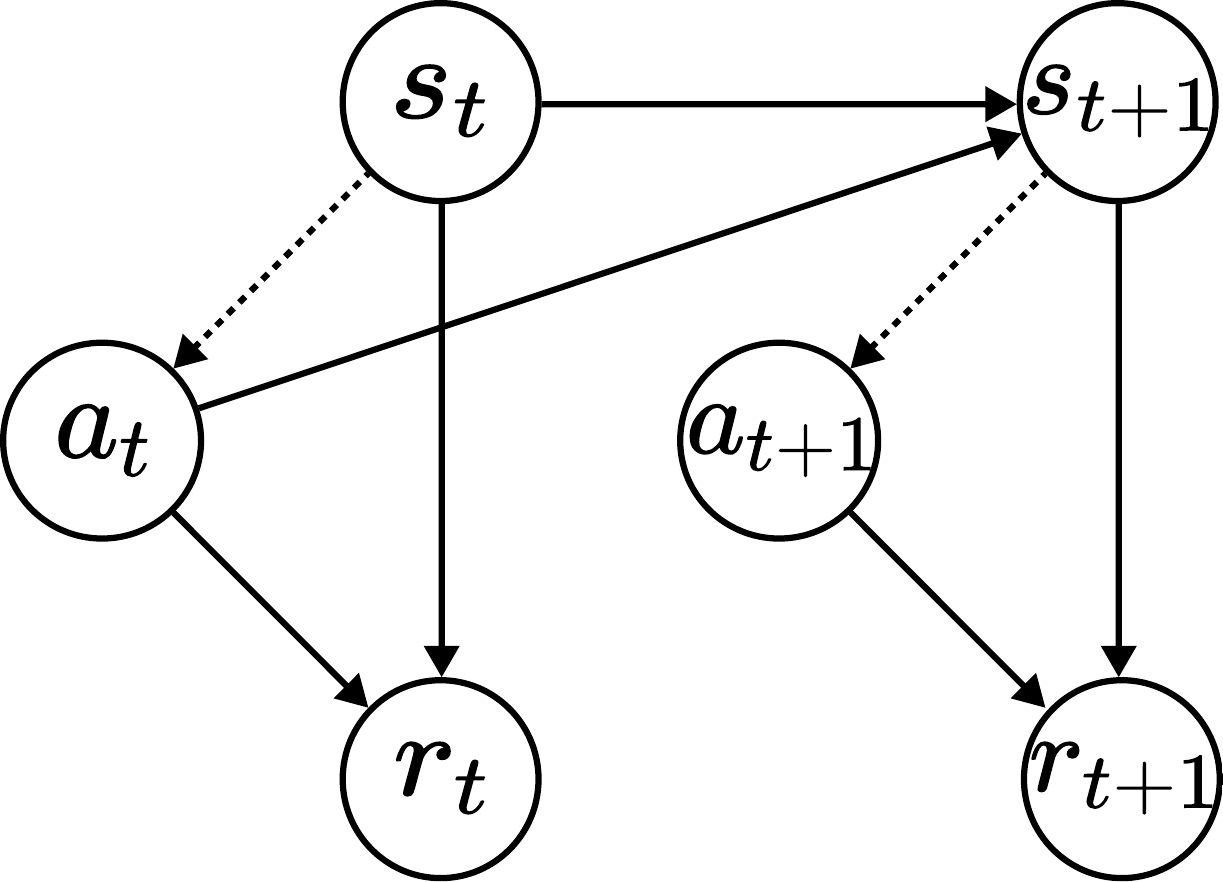}
    \caption{\textbf{Causal graph in our RL setting.} Solid arrows: environment-induced transitions and reward generation, Dotted arrows: denote policy-induced dependencies.}
    \label{fig:scm}
    \vspace{-1.8em}
\end{wrapfigure}


A Structural Causal Model (SCM) describes causal relationships between variables~\citep{pearl2009causality, pearl2011transportability}. An SCM $\mathcal{M}$ is defined as a tuple $(\mathbf{U}, \mathbf{V}, \mathscr{F}, P)$, where $\mathbf{U}$ is a set of exogenous (unobserved) variables, $\mathbf{V}$ is a set of endogenous (observed) variables, $\mathscr{F} = \{ f_V : \mathit{Pa}(V) \cup U_V \to V \}_{V \in \mathbf{V}}
$ is a set of structural functions, where each $f_V$ determines $V$ based on its endogenous parents $\mathit{Pa}(V) \subseteq \mathbf{V}$ (endogenous variables that directly influence $V$) and exogenous inputs $U_V \subseteq \mathbf{U}$ (exogenous variables that directly influence $V$), and $P(\mathbf{U})$ is the distribution over $\mathbf{U}$, from which the exogenous variables are sampled.


From an RL perspective, MDP can be interpreted through the lens of SCM, where the state space $\mathbf{S}$, the action space $\mathbf{X}$, and the reward space $\mathbf{Y}$ are subsets of the set of endogenous variables $\mathbf{V}$, and the randomness in the reward and transition functions is captured by the set of exogenous variables $\mathbf{U}$~\citep{li2024causallyaligned}. In this paper, we assume a deterministic setting without exogenous variables, i.e., there is no underlying randomness in the environment. The resulting causal structure is illustrated in Figure~\ref{fig:scm}. Here, $s_t, s_{t+1} \in \mathbf{S}$, $a_t, a_{t+1} \in \mathbf{X}$, and $r_t, r_{t+1} \in \mathbf{Y}$, and the structural functions in $\mathscr{F}$ map $(s_t, a_t)$ to both $s_{t+1}$ and $r_t$.

\subsection{Curriculum reinforcement learning via constrained optimal transport}
Our method is built upon CURROT~\citep{klink2022currot}, which makes a curriculum using the optimal transport problem with constraints.
\begin{equation}
p_W(c) = \arg\min_p \mathcal{W}_2(p(c), \mu(c)) 
\quad \text{s.t.} \quad 
\begin{cases}
p(c) > 0 \Rightarrow J(\pi, c) \geq \delta & \forall c \in \mathcal{C} \\
\mathcal{W}_2(p(c), q(c)) \leq \epsilon
\end{cases}
\label{eq:optimal_transport}
\end{equation}
It tries to minimize the 2-Wasserstein distance $\mathcal{W}_2$, the distance between the current source task distribution $p(c)$ and the target task distribution $\mu(c)$, with two constraints: (1) among all tasks, only those with nonzero probability under $p(c)$-i.e., tasks actually selected for training-must yield an expected return above a threshold $\delta$ (which decreases over time), and (2) the Wasserstein distance between consecutive source task distributions $p(c)$ and $q(c)$ should be lower than the threshold $\epsilon$. This optimization enables the construction of a curriculum that smoothly approaches the target distribution, while maintaining per-task expected return $J(\pi,c)$ above a decreasing threshold $\delta$ and bounding the distributional shift between successive task distributions.

\section{Method}
\label{sec:method}
We propose a curriculum learning method that encourages the agent to explore tasks with underexplored causal structures. To achieve this, we introduce the Causal Misalignment (CM) score, which quantifies how unfamiliar a task’s causal structure is relative to the agent’s previously learned experience. We empirically analyze the relationship between the CM score and underlying causal differences in the CausalWorld environment~\citep{ahmed2021causalworld}. We incorporate this score into CURROT to iteratively update the task distribution. By guiding the agent toward tasks with high causal misalignment, our method promotes more efficient exploration and fosters generalizable learning.

\subsection{Prioritizing Causally underexplored Tasks in Curriculum Learning}
To construct an effective curriculum, it is essential to accurately quantify how a new task differs from previously encountered ones in terms of learning dynamics and structural complexity~\citep{hughes2024open}. If the SCMs of tasks were available, they could serve as a principled basis for comparing tasks in terms of their underlying causal mechanisms~\citep{li2024causallyaligned}. However, direct access to SCMs is rarely feasible in realistic reinforcement learning environments, and uncovering such structures from purely observational data remains a challenging problem~\citep{zanga2022survey, zeng2024survey}. In deterministic settings, the interplay among state, action, transition, and reward defines a structured dependency that implicitly captures the task's causal structure. Thus, the causal difference between two tasks can be approximated by aggregating the differences in these four components. To estimate each component-wise difference, we train dedicated models using trajectories collected from previously experienced tasks and compute disagreement among ensemble predictions.
\begin{equation*}
\label{eq:component_model}
    \begin{aligned}
        \text{State encoder: } &z_t^s \sim q_{\phi_s}(z_t^s \mid s_t) \\
        \text{State decoder: } &\hat{s}_t \sim p_{\theta_s}(\hat{s}_t \mid z_t^s) \\
        \text{Action encoder: } &z_t^a \sim q_{\phi_a}(z_t^a \mid a_t) \\
        \text{Action decoder: } &\hat{a}_t \sim p_{\theta_a}(\hat{a}_t \mid z_t^a) \\
        \text{Transition predictor: } &\hat{s}_{t+1} \sim p_{\phi_\tau}(\hat{s}_{t+1} \mid s_t, a_t) \\
        \text{Reward predictor: } &\hat{r}_t \sim p_{\phi_r}(\hat{r}_t \mid s_t, a_t)
    \end{aligned}
\end{equation*}

We train six separate neural networks. (\(\hat{s}\), \(\hat{a}\), \(\hat{s}_{t+1}\), \(\hat{r}_t\)) denote the model's reconstruction or prediction of a corresponding variable. The state encoder $q_{\phi_s}(z_t^s \mid s_t)$ and the state decoder $p_{\theta_s}(\hat{s}_t \mid z_t^s)$ encode the current state $s_t$ into a latent representation $z_t^s$ and reconstruct it to the state $\hat{s}_t$. The action encoder $q_{\phi_a}(z_t^a \mid a_t)$ and the action decoder $p_{\theta_a}(\hat{a}_t \mid z_t^a)$ encode the action $a_t$ into a latent representation $z_t^a$ and reconstruct it to the action $\hat{a}_t$. The transition predictor $p_{\phi_\tau}(\hat{s}_{t+1} \mid s_t, a_t)$ predicts the next state based on the current state and action, approximating the environment's transition function. The reward predictor $p_{\phi_r}(\hat{r}_t \mid s_t, a_t)$ predicts rewards for the current state and action, modeling the environment's reward function. The detailed architecture of these neural networks is provided in Appendix~\ref{sec:appendix:implementation_details:network_structures}.
\begin{equation}
\label{eq:component_losses}
    \begin{aligned}
        \mathcal{L}_{\text{state}} &= \|\hat{s}_t - s_t\|^2_2 + \beta_1 \cdot \text{KL}(q_{\phi_s}(z_t^s \mid s_t) \mid \mathcal{N}(0, I)) \\
        \mathcal{L}_{\text{action}} &= \|\hat{a}_t - a_t\|^2_2 + \beta_2 \cdot \text{KL}(q_{\phi_a}(z_t^a \mid a_t) \mid \mathcal{N}(0, I)) \\
        \mathcal{L}_{\text{transition}} &= \|\hat{s}_{t+1} - s_{t+1}\|^2_2 \\
        \mathcal{L}_{\text{reward}} &= \|\hat{r}_t - r_t\|^2_2
    \end{aligned}
\end{equation}
These models are trained using separate loss functions, each guiding a distinct component of the causal structure. The state and action losses, \(\mathcal{L}_{\text{state}}\) and \(\mathcal{L}_{\text{action}}\), guide the model to accurately reconstruct the state and action, respectively, ensuring that the latent representations capture sufficient information to regenerate the original inputs. These losses also regularize the latent distributions toward a standard Gaussian prior via KL divergence as in~\citet{higgins2017beta}. The transition loss, \(\mathcal{L}_{\text{transition}}\), guides the model to approximate the transition dynamics by predicting the next state given the current state and action. Similarly, the reward loss, \(\mathcal{L}_{\text{reward}}\), promotes accurate modeling of the reward function by supervising reward prediction from the current state and action.

While component-wise neural networks enable the separate modeling of individual environment components, simply training these models is insufficient to assess the causal difference between a new task and previously learned tasks. To quantify this difference, we train an ensemble of K networks for each of the four components-state, action, transition, and reward-and measure the structural unawareness of each component through ensemble disagreement, computed as the standard deviation of predictions across ensemble members as in~\citet{zhang2020automatic}.
\begin{equation}
\label{eq:disagreement}
\text{Disagreement}_i = \mathrm{std}\left( \left\{ \hat{y}_i^{(k)}(\cdot) \right\}_{k=1}^{K} \right), \quad i \in \{\text{state}, \text{action}, \text{transition}, \text{reward}\}
\end{equation}
Here, \(\hat{y}_i^{(k)}\) denotes the prediction of the \(k\)-th ensemble member for component \(i\). A high level of disagreement indicates that the task's causal structure significantly differs from previously encountered tasks, which we interpret as causal misalignment, suggesting that the agent has not yet confidently captured the underlying structure. In contrast, low disagreement implies that the agent has developed a reliable understanding of the corresponding environment component.

We aggregate these component-wise disagreements into a CM score, which quantifies the overall causal structure unawareness associated with a given context. The CM score is computed as a weighted sum of the individual component disagreements.

\begin{equation}
\label{eq:misalignment_score}
    \text{CM}(c) = \sum_{i} w_i \cdot \text{Disagreement}_i
 \quad i \in \{\text{state}, \text{action}, \text{transition}, \text{reward}\}
\end{equation}
This score quantifies the extent to which the structural properties of a given task deviate from those captured by the agent’s current model. Tasks with higher misalignment scores are prioritized during curriculum sampling, guiding the agent toward regions of the task space where its understanding of the causal structure remains underdeveloped. Details of the weighting hyperparameters are provided in Appendix~\ref{sec:hyperparameters}.

We extend CURROT by incorporating causal misalignment into the context distribution update. At each iteration, the agent samples a batch of tasks from the current context distribution, collects interaction trajectories, and updates its policy. For each task, component-wise ensemble disagreements are computed and aggregated into a CM score via a weighted sum. This score, combined with episodic rewards, captures both structural novelty and learnability. These signals are stored and used to update the context distribution through an optimal transport-based objective (Eq.~\ref{eq:optimal_transport}). The full procedure is described in Appendix~\ref{sec:appendix_overall_workflow}.

\subsection{Disagreement as a Proxy for Causal Difference: A Toy Example in CausalWorld}

\begin{figure}[h]
    \centering
    \includegraphics[width=0.8\linewidth]{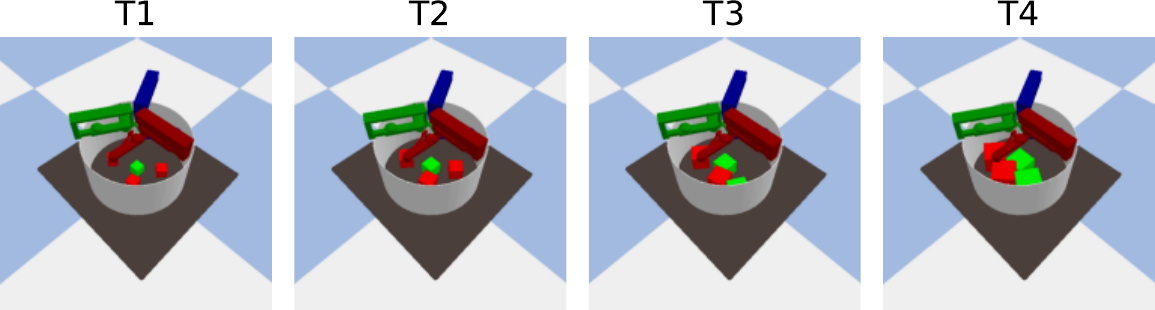}
    \caption{
    \textbf{Visualization of tasks T1–T4 from the CausalWorld General environment.} In this setup, agents receive rewards proportional to the intersection ratio between each block and the goal configuration. From T1 to T4, we progressively increase the block size, action magnitude, and reward scale, inducing increasing causal differences between tasks.
    }
    \label{fig:causal_world_tasks}
\end{figure}

To assess the effectiveness of the CM score in capturing task-level causal differences, we conduct a controlled experiment using the CausalWorld General environment~\citep{ahmed2021causalworld}. CausalWorld provides a realistic and structured environment suitable for analyzing how modular disagreement relates to underlying causal changes.

We construct four tasks, T1 through T4 (Figure~\ref{fig:causal_world_tasks}), by progressively increasing the block size, action magnitude, and reward scale, thereby inducing increasing levels of causal variation across tasks. These modifications induce monotonic increases in causal differences relative to T1. Detailed task configurations are provided in Appendix~\ref{sec:appendix_analysis_disagreement_metrics}. We use the same model architecture described in Appendix~\ref{sec:appendix:implementation_details:network_structures}, except that, for this experiment, the hidden dimension of the transition and reward predictors is fixed to 64 and the ensemble size is set to 5. We designate T1 as the base task and pretrain on it for 10 episodes using a random policy. Then, we collect 5 episodes of random trajectories from each of the four tasks and evaluate the disagreement scores across state, action, transition, and reward models.

\begin{figure}[h]
    \centering
    \includegraphics[width=\linewidth]{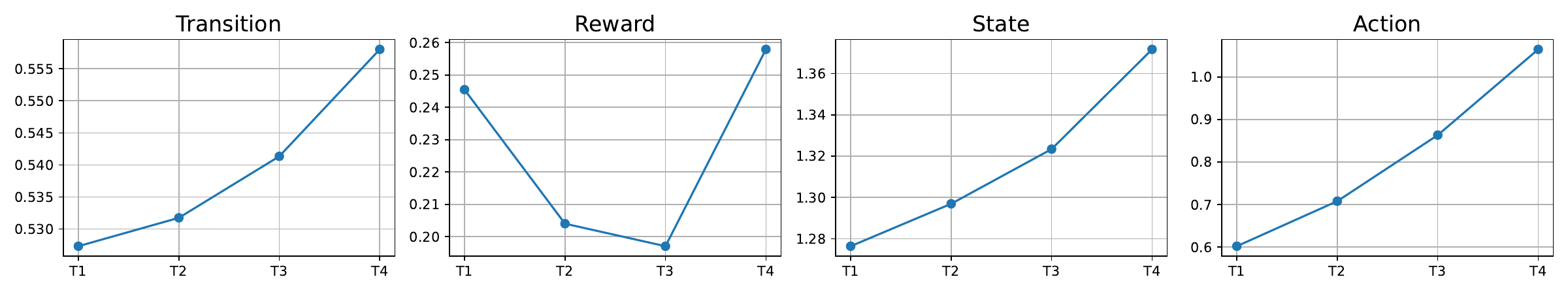}
    \caption{
    \textbf{Disagreement metrics across tasks T1–T4.} Each metric is computed after training on T1 for 10 episodes and sampling 5 episodes from each task. 
    Transition, state, and action disagreements increase monotonically, suggesting they effectively reflect causal differences. 
    Reward disagreement shows no clear trend due to reward sparsity and random trajectory collection.
    }
    \label{fig:causal_misalignment_scores}
    \vspace{-1em}
\end{figure}

Figure~\ref{fig:causal_misalignment_scores} presents the averaged disagreement metrics over three random seeds. Transition, state, and action disagreements increase monotonically from T1 to T4, aligning well with the intended causal differences induced by task variations. In contrast, reward disagreement shows no consistent trend, likely due to sparse reward signals and the use of random policies during data collection. These results validate that the proposed component-wise disagreement metrics-especially for state, action, and transition-serve as effective proxies for capturing structural differences across tasks.

Moreover, further analysis in Appendix~\ref{sec:appendix_analysis_disagreement_metrics} reveals that transition disagreement is primarily influenced by both block size and action magnitude, action disagreement correlates with action scale, and state disagreement increases with block size. These findings reinforce our interpretation that modular disagreement meaningfully reflects the causal components underlying task variations.

\section{Experiments}
\label{sec:experiments}

\begin{figure}[h]
    \centering
    \begin{subfigure}[b]{0.20\linewidth}
        \includegraphics[width=\linewidth]{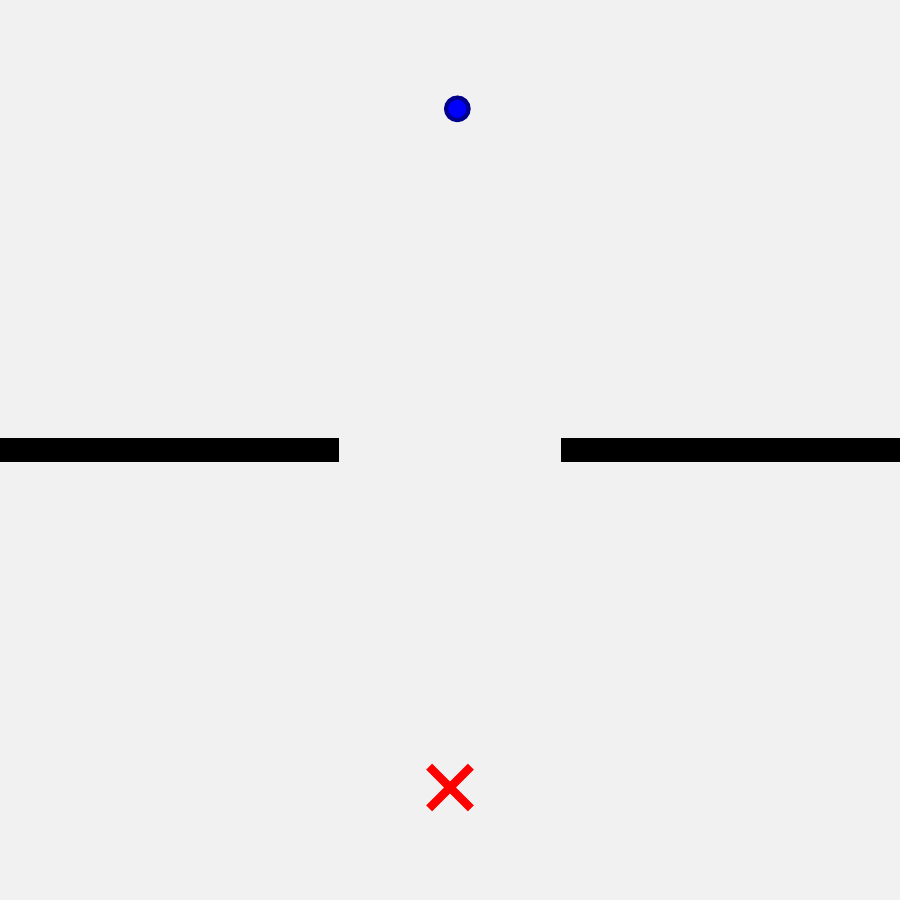}
        \caption{Point Mass}
        \label{fig:pointmass_env}
    \end{subfigure}
    \quad
    \begin{subfigure}[b]{0.41\linewidth}
        \includegraphics[width=\linewidth]{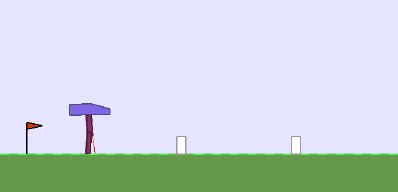}
        \caption{BipedalWalker}
        \label{fig:bipedalwalker_env}
    \end{subfigure}
    \caption{\textbf{Illustrations of the environments used in our experiments.} (a) Point Mass. (b) BipedalWalker.}
    \label{fig:envs}
\end{figure}

We evaluated our method, CP-DRL, on two benchmark environments: Point Mass (PM) and Bipedal Walker (BW). For the PM environment, we used Proximal Policy Optimization (PPO) as the student algorithm, while for BW, we employed Soft Actor-Critic (SAC) as in~\citet{klink2022currot}. Visualizations of each environment are provided in Figure~\ref{fig:envs}.

For comparison, we evaluated several curriculum learning baselines, including CURROT, SPRL, GoalGAN, ALPGMM, ACL, PLR, and VDS~\citep{klink2022currot, klink2021probabilistic, florensa2018automatic, portelas2020teacher, graves2017automated, jiang2021prioritized, zhang2020automatic}.  All experiments are conducted with multiple random seeds to ensure statistical reliability, 10 seeds for PM, 5 seeds for BW trivial tasks, and 3 seeds for BW infeasible tasks. We present a comparison with CURROT, the strongest among the baselines, and include a comprehensive evaluation in Appendix~\ref{sec:comprehensive_experiments_results}.

\subsection{Point Mass}
\begin{figure}[t]
\vspace{-1em}
\label{fig:pointmass_main}
    \centering
    \includegraphics[width=0.8\linewidth]{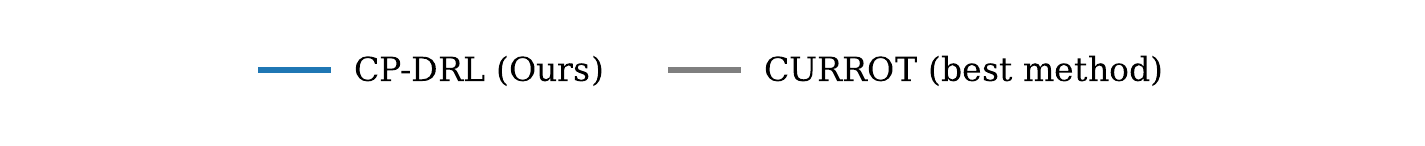}
    \begin{subfigure}[b]{0.32\linewidth}
        \includegraphics[width=\linewidth]{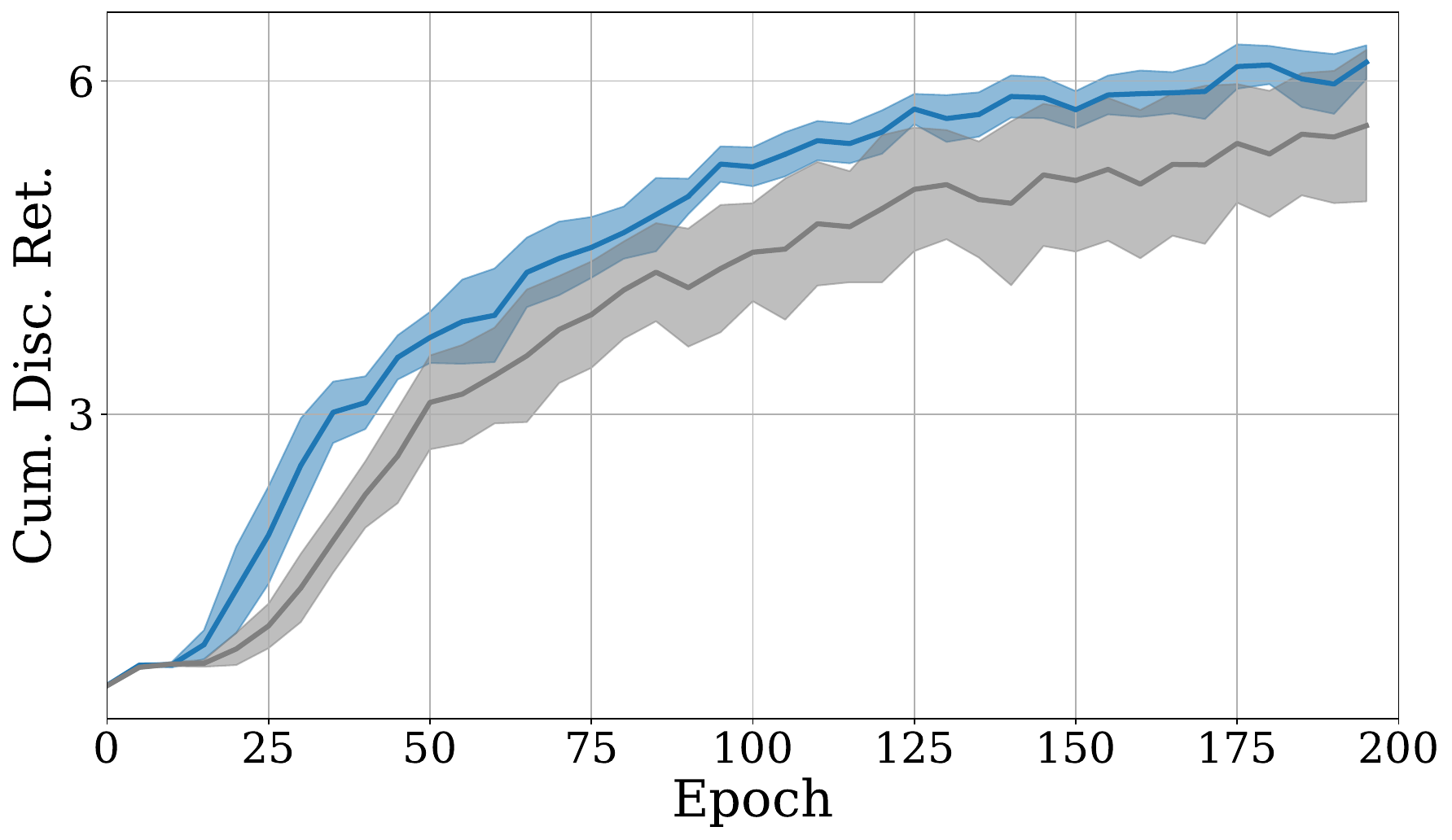}
        \caption{Cumulative discounted return}
        \label{fig:pointmass_performance}
    \end{subfigure}
    \begin{subfigure}[b]{0.32\linewidth}
        \includegraphics[width=\linewidth]{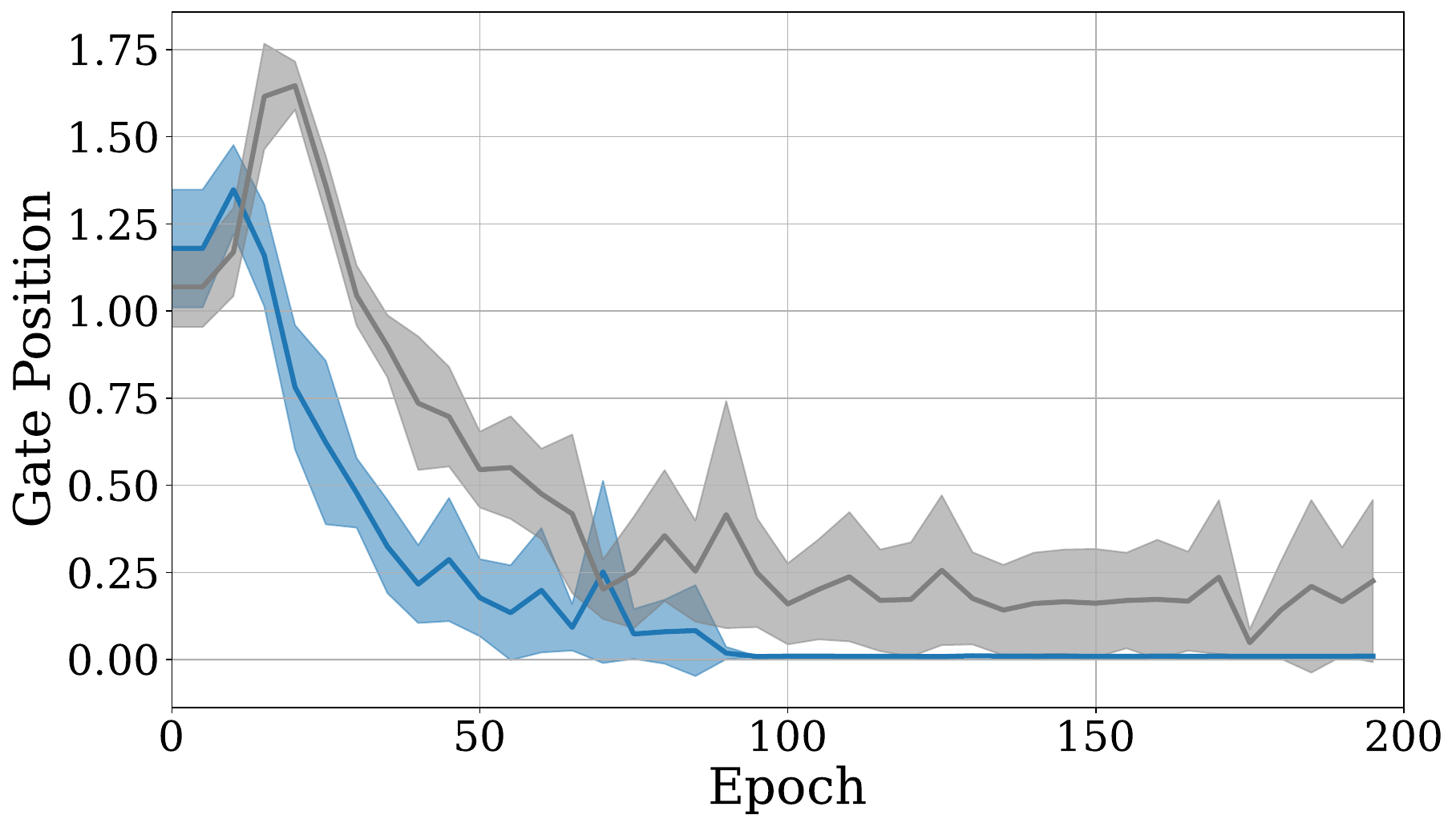}
        \caption{Gate position}
        \label{fig:pointmass_gate_position}
    \end{subfigure}
    \begin{subfigure}[b]{0.32\linewidth}
    \includegraphics[width=\linewidth]{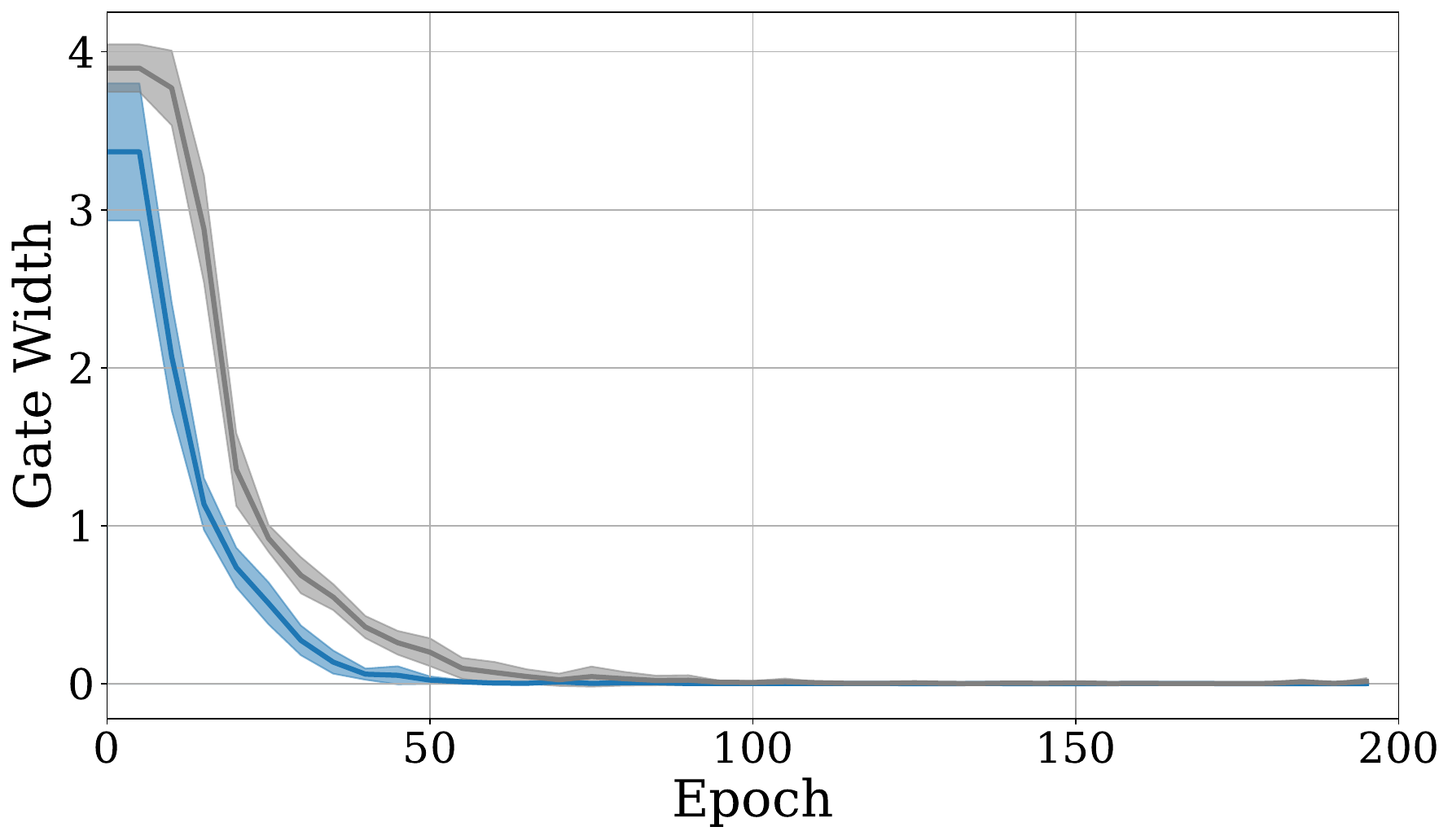}
        \caption{Gate width}
        \label{fig:pointmass_gate_width}
    \end{subfigure}
    \caption{\textbf{Performance comparison between CP-DRL and CURROT in Point Mass environment.} (a) Cumulative discounted return. (b) Median distance to the target gate position. (c) Median distance to the target gate width. All curves show the mean with 95\% confidence intervals.}
    \vspace{-1em}
\end{figure}

The PM environment requires controlling a point agent to navigate through a narrow gate in order to reach a target location on the opposite side of a wall~\citep{klink2020scrl, klink2020sprl, klink2021probabilistic}. The task difficulty is modulated by the gate’s width and position. The target context distribution \(\mu(c)\) is bimodal, corresponding to two gate positions on opposite sides. Each training epoch consists of 4,096 rollouts, and all methods are trained for 200 epochs to ensure convergence. For a detailed description of the environment, please refer to Appendix~\ref{sec:appendix_point_mass}.

CP-DRL demonstrates superior performance throughout training, reaching a return of \(6.17 \pm 0.08\) at epoch 195. This substantially outperforms the second-best method, CURROT (\(5.6 \pm 0.34\)), with an improvement of approximately \(10.2\%\). Moreover, CP-DRL maintains consistently low standard errors during training, indicating more stable learning compared to CURROT, which exhibits increasing variance over time. The overall performance trend is illustrated in Figure~\ref{fig:pointmass_performance}.

We also examine how CP-DRL adapts the curriculum over time by visualizing the evolution of gate positions and gate width precision. The agent quickly converges to tasks aligned with the target gate position and progressively focuses on narrower gate widths, demonstrating that the curriculum drives the agent toward more precise and challenging contexts. These patterns reflect CP-DRL’s ability to facilitate efficient exploration by prioritizing causally unfamiliar tasks, thereby accelerating convergence and enhancing adaptation. Visualizations are provided in Figure~\ref{fig:pointmass_gate_position} and Figure~\ref{fig:pointmass_gate_width}.

\subsection{Bipedal Walker}

\begin{figure}[h]
    \vspace{-1em}
    \centering
    \includegraphics[width=0.8\linewidth]{figures/BW_legend_main.pdf}
    
    \begin{subfigure}{0.45\linewidth}
        \includegraphics[width=\linewidth]{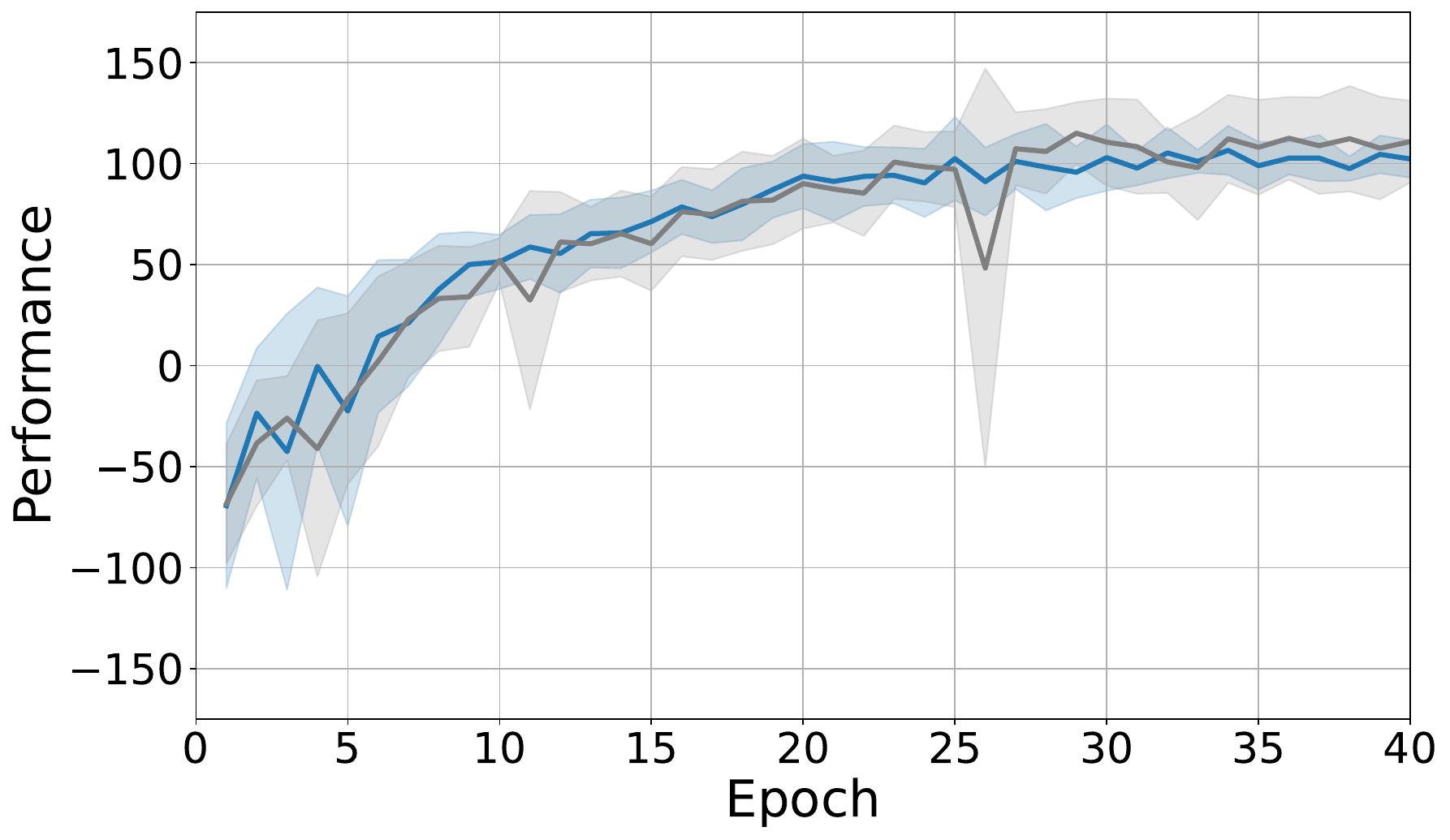}
        \caption{Performance in Bipedal Walker (Trivial)}
        \label{fig:bw_trivial}
    \end{subfigure}
    \begin{subfigure}{0.45\linewidth}
        \includegraphics[width=\linewidth]{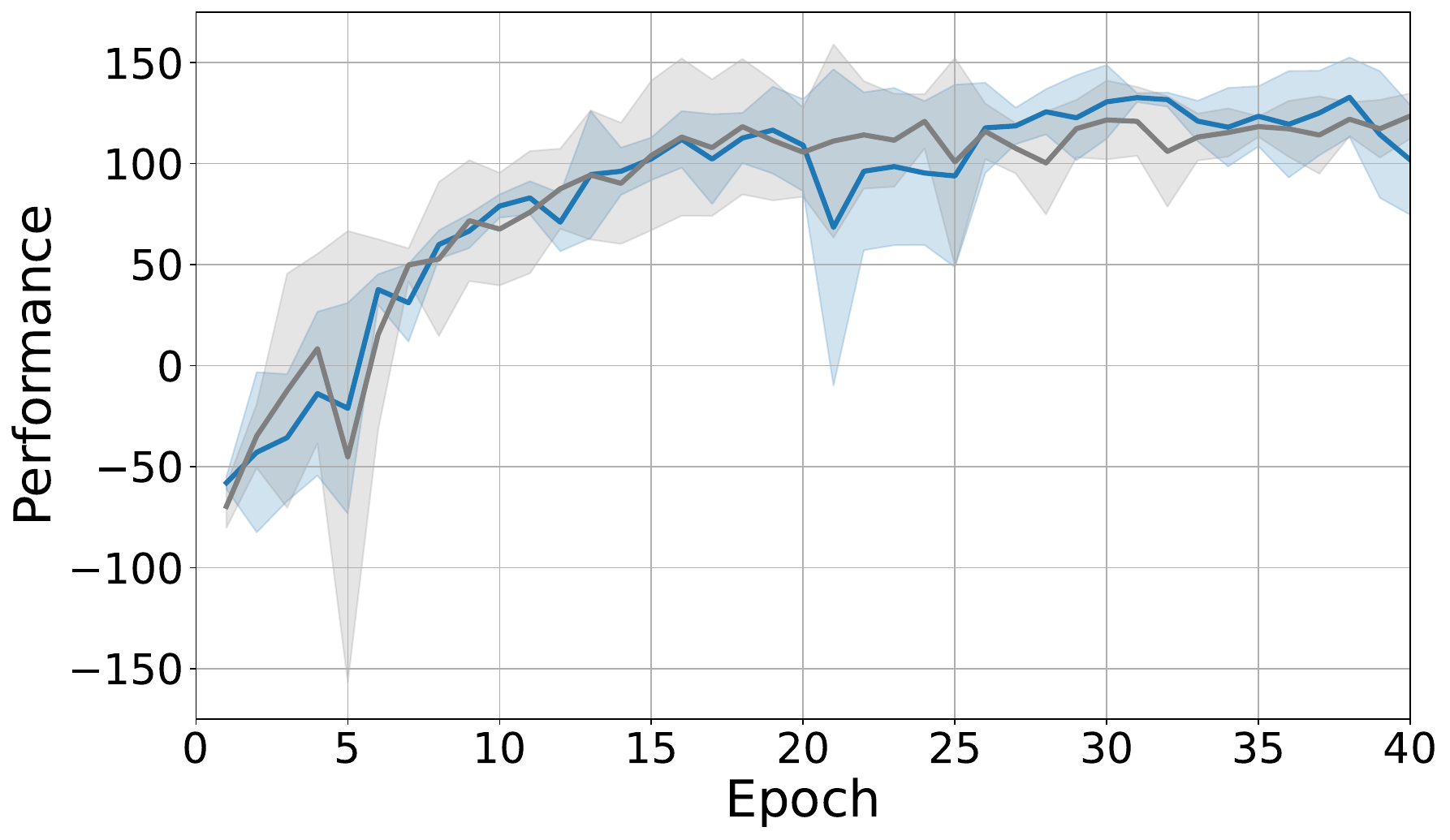}
        \caption{Performance in Bipedal Walker (Infeasible)}
        \label{fig:bw_infeasible}
    \end{subfigure}
    
    \caption{
    \textbf{Performance comparison between CP-DRL and CURROT in BipedalWalker.} 
    (a) \textbf{Trivial}. 
    (b) \textbf{Infeasible}. 
    All curves show the mean with 95\% confidence intervals.}
    \label{fig:bw_combined}
\end{figure}

The BW environment is a continuous control benchmark where a bipedal agent must traverse uneven terrain consisting of randomly spaced stumps. Each task is parameterized by two contextual variables: stump height and stump spacing, which respectively control the vertical and horizontal difficulty of locomotion. In the Trivial setting, stump height is sampled from \([-3, 3]\) and clipped at zero, resulting in approximately half of the tasks being flat and easy. In contrast, the Infeasible setting uses a wider range \([0, 9]\) for stump height, leading to many tasks that are physically impossible to solve. Stump spacing is sampled from \([0, 6]\) in both settings. Each training epoch consists of \(5 \times 10^{4}\) environment steps, and all methods are trained for 40 epochs to ensure convergence. For further environment details, refer to Appendix~\ref{sec:appendix_bipedalwalker}.

In the Trivial setting (Figure~\ref{fig:bw_trivial}), CP-DRL achieves the fastest convergence among all methods, reaching a return of \(93.82 \pm 8.12\) at 20k steps and maintaining stable performance thereafter. While CURROT attains slightly higher final returns, CP-DRL exhibits a lower standard error for most of the training period. Other methods, such as ALP-GMM and GOALGAN, exhibit moderate success, while SPRL and Random sampling fail to achieve meaningful progress.

In the Infeasible setting (Figure~\ref{fig:bw_infeasible}), CP-DRL outperforms all baselines during the mid-training phase, achieving the highest mean return of \(130.61 \pm 9.32\) at 30k steps. Although CURROT achieves a higher final return (\(123.58 \pm 5.72\) vs. \(101.85 \pm 13.87\)), CP-DRL reaches superior performance earlier in training. In contrast, several baselines, such as SPRL, RIAC, and Random, show severe instability or collapse entirely. 

Together, these results confirm that CP-DRL not only achieves competitive performance but also excels in stability and convergence speed, especially in structurally diverse or difficult environments.

\section{Limitations}
\label{sec:limitation}

Our approach, while effective, has several limitations that highlight opportunities for future improvement. The method depends on approximating SCMs to estimate differences between tasks, but the quality of this approximation is inherently limited by the choice of model architecture and training strategy. Incorporating more expressive or causally grounded models, such as those based on counterfactual reasoning or latent causal representations, may lead to more accurate consistency estimation and improved curriculum design.

The current evaluation is confined to a small set of benchmark environments, leaving open the question of how well the method generalizes to more complex or diverse settings. Environments with sparse rewards or unobserved confounders may present different challenges, and extending our experiments to these domains would help validate the robustness and scalability of the approach.

Although the framework integrates naturally with CURROT, its applicability to other curriculum learning paradigms has not been explored. Understanding whether the proposed causal-guided strategy can be transferred to alternative frameworks would clarify its general utility and flexibility.

Finally, in its current form, the method uses heuristic rules, such as thresholds or fixed weightings. Developing more principled weighting mechanisms, potentially through meta-learning or causal score matching, could improve the adaptivity and performance of the curriculum.

\section{Conclusion}
\label{sec:conclusion}

We presented CP-DRL, a curriculum learning framework that leverages causal reasoning to guide task selection. Instead of relying on access to true SCMs, CP-DRL quantifies the agent's unfamiliarity with the causal structure of a task through the disagreement of the state, action, transition, and reward predictors. These structural signals are integrated into a transport-based curriculum optimization scheme, enabling the agent to explore causally underexplored regions while gradually aligning with the target task distribution.

Through experiments on two reinforcement learning benchmarks, PM and BW, we demonstrate the effectiveness of CP-DRL in improving curriculum learning. Our findings demonstrate the promise of causal signals as a general tool for curriculum design in RL. Future work may explore tighter integration with causal representation methods, application to real-world robotic systems, or extension to partially observable and sparse reward settings. We hope that this work inspires further research into causally grounded learning and structure-aware exploration for adaptive agents in open-ended environments.

\bibliography{main}
\bibliographystyle{rlj}

\beginSupplementaryMaterials
\appendix

\renewcommand*{\thesection}{\Alph{section}}
\section{Comprehensive Experiments Results}
\label{sec:comprehensive_experiments_results}

\vspace{-1em}

\begin{figure}[h]
    \centering
    \includegraphics[width=0.9\linewidth]{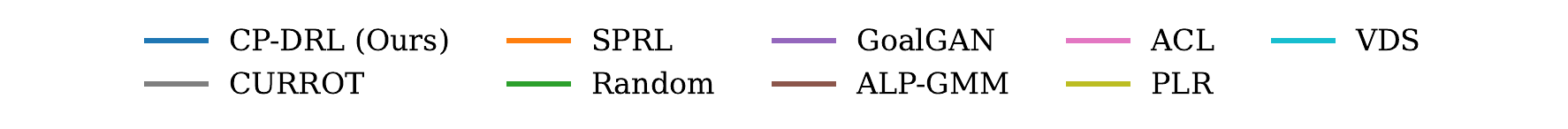}
    \begin{subfigure}[b]{0.32\linewidth}
        \includegraphics[width=\linewidth]{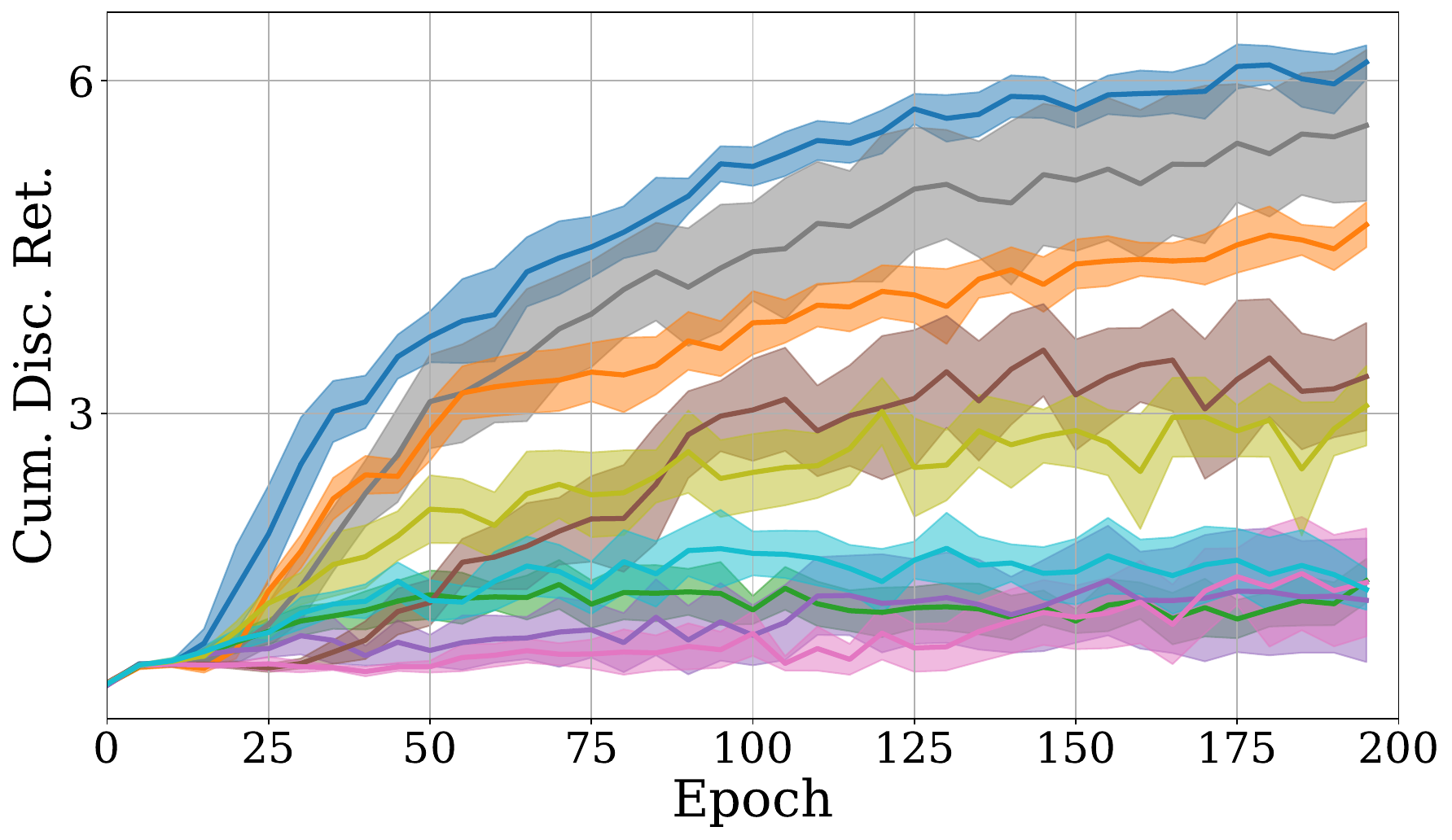}
        \caption{Cumulative discounted return}
        \label{fig:pointmass_performance_baseline}
    \end{subfigure}
    \begin{subfigure}[b]{0.32\linewidth}
        \includegraphics[width=\linewidth]{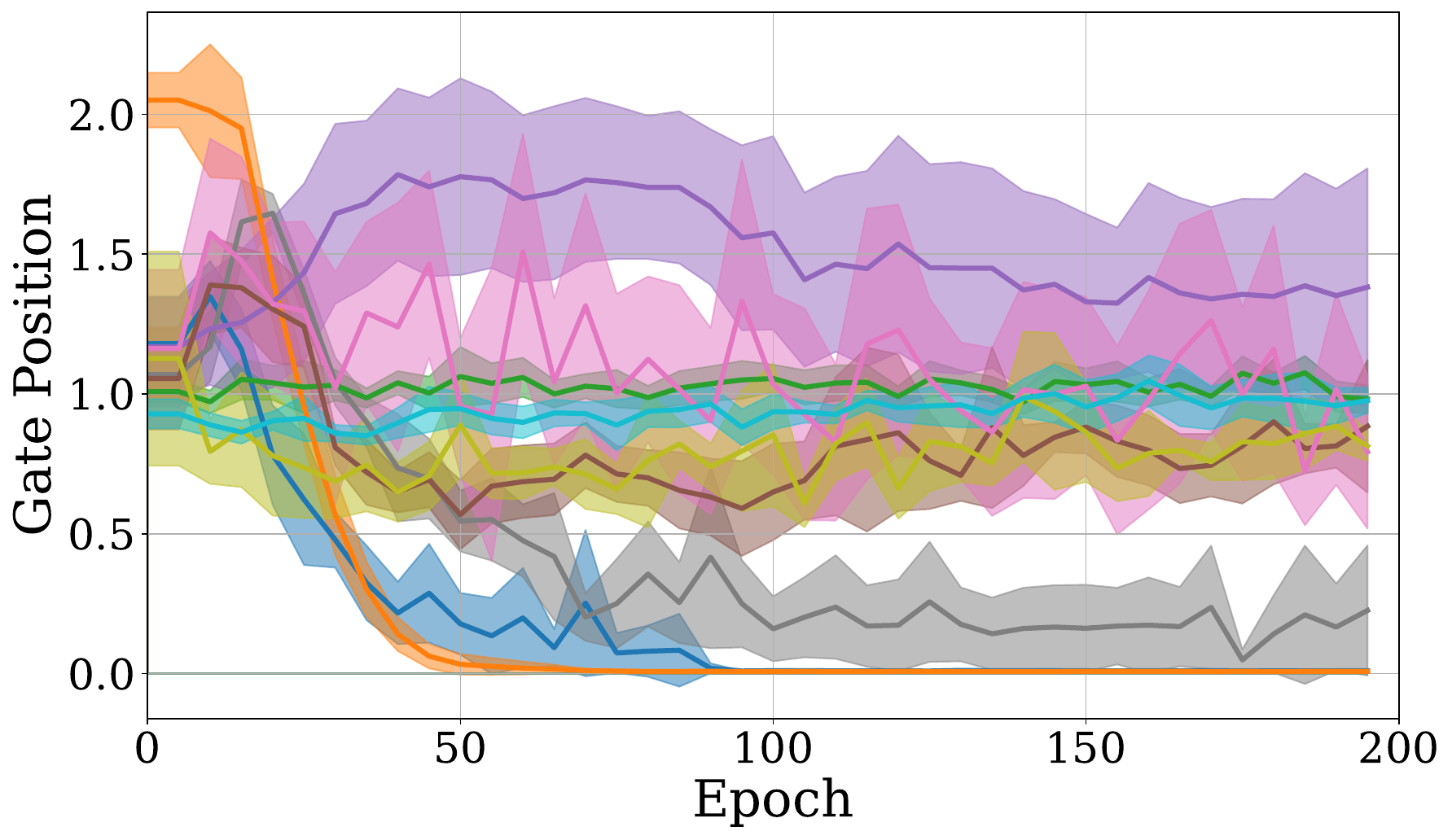}
        \caption{Gate position}
        \label{fig:pointmass_gate_position_baseline}
    \end{subfigure}
    \begin{subfigure}[b]{0.32\linewidth}
    \includegraphics[width=\linewidth]{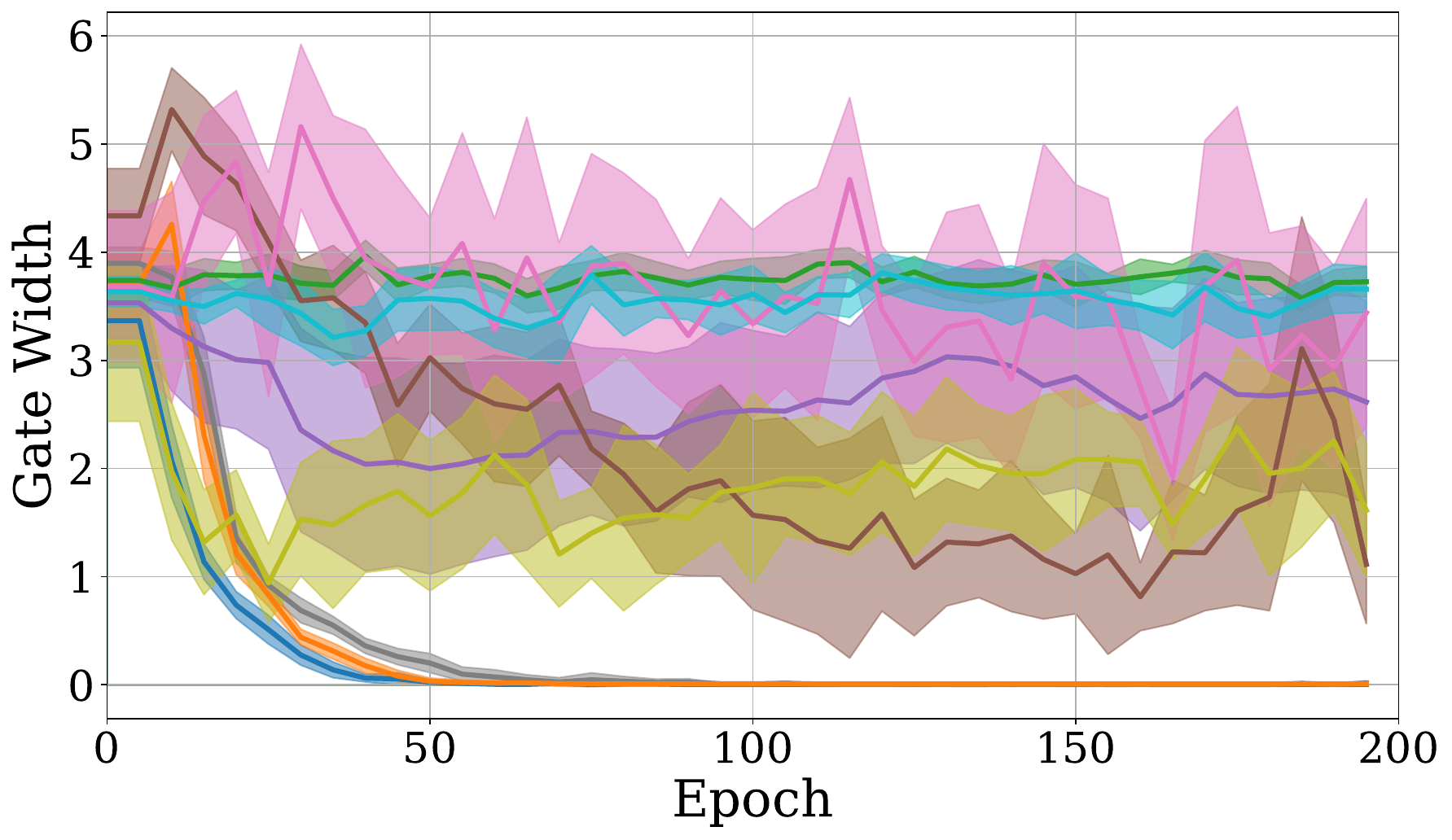}
        \caption{Gate width}
        \label{fig:pointmass_gate_width_baseline}
    \end{subfigure}
    \caption{\textbf{Performance comparison in Point Mass environment under different curriculum methods.} (a) Cumulative discounted return over 200 epochs. (b) Median distance to the target gate position (c) Median distance to the target gate width. All curves show the mean with 95\% confidence intervals.}
    \label{fig:pointmass_baseline}
\end{figure}

\vspace{-1em}

\begin{figure}[h]
    \centering
    \includegraphics[width=0.7\linewidth]{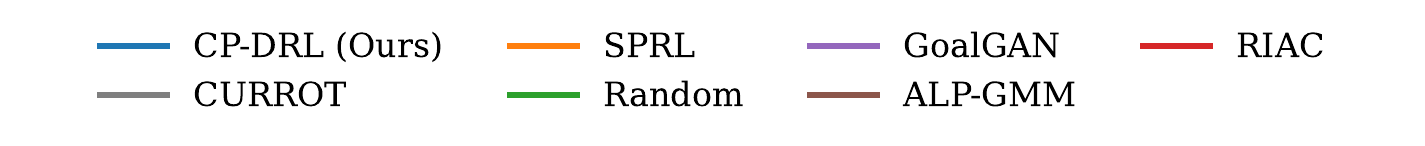}
    
    \begin{subfigure}{0.45\linewidth}
        \includegraphics[width=\linewidth]{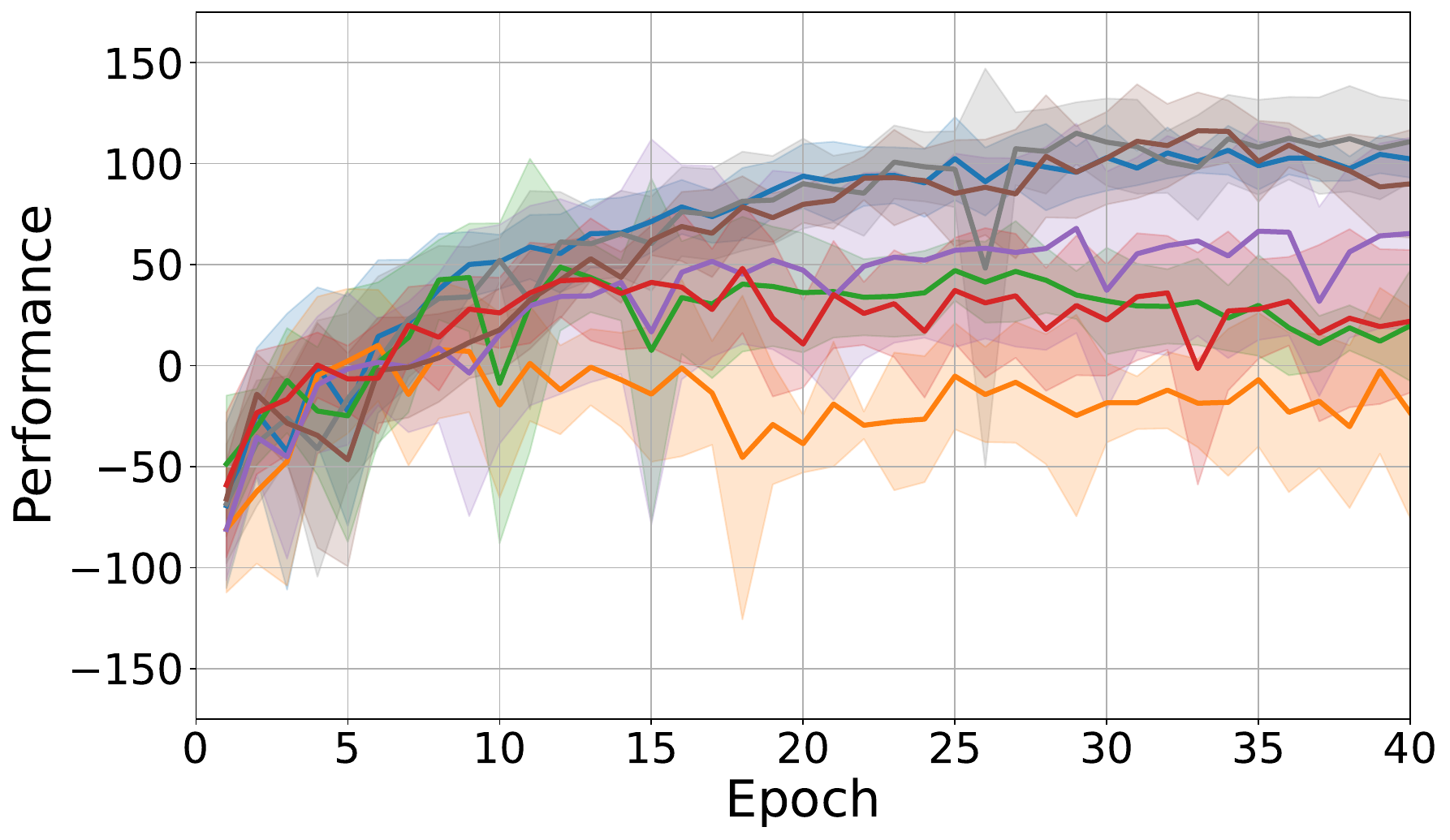}
        \caption{Performance in Bipedal Walker (Trivial)}
        \label{fig:bw_trivial_baseline}
    \end{subfigure}
    \begin{subfigure}{0.45\linewidth}
        \includegraphics[width=\linewidth]{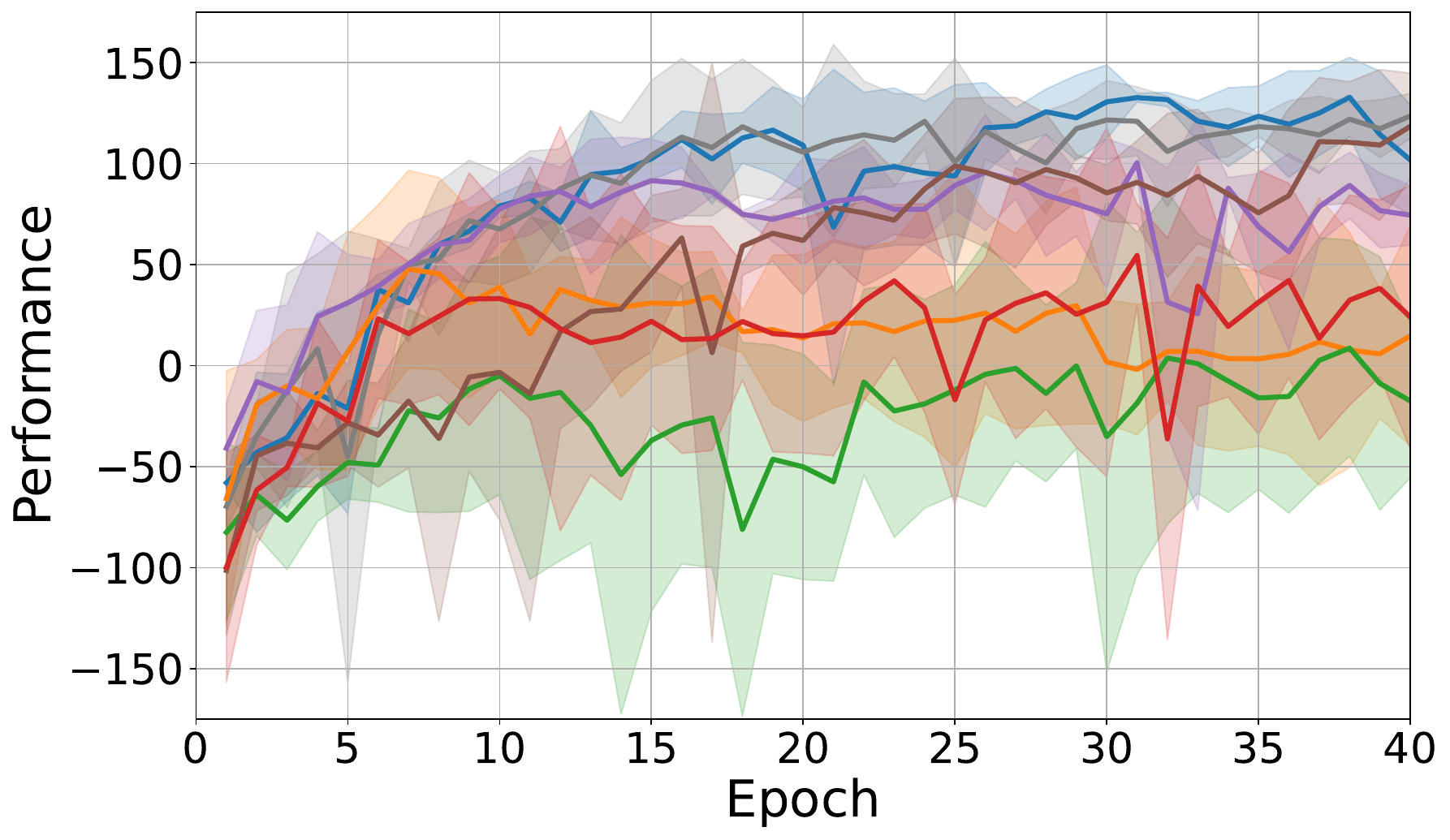}
        \caption{Performance in Bipedal Walker (Infeasible)}
        \label{fig:bw_infeasible_baseline}
    \end{subfigure}
    
    \caption{
    \textbf{Performance comparison of curriculum methods in BipedalWalker.}
    (a) Trivial, (b) Infeasible. All curves show the mean with 95\% confidence intervals.
    }
    \label{fig:bw_baseline}
\end{figure}

\section{Implementation Details}
\label{sec:implementation_details}



\subsection{Point Mass Environment}
\label{sec:appendix_point_mass}
The Point Mass environment presents a continuous 2-dimensional task where the agent, starting from $[x=0,y=0]$, must pass through the narrow gate at $y=0$ and reach the goal at $[0,-3]$. The agent's state is represented as $[x, v_x, y, v_y]$ where $[x, y]$ is a position and the $[v_x, v_y]$ is a speed~\citep{klink2020scrl, klink2020sprl, klink2021probabilistic}. The action space is a 2-dimensional continuous vector within $[-10.0, 10.0]^2$, so the action $[a_x,a_y]$ is clipped into $[-10.0, 10.0]^2$. The agent starts from the position $[0.0,3.0]$ with no velocity. The agent's state is updated 10 times per step with a time interval of 0.01 using Euler integration to approximate the differential equation. The update rule for the x-axis is given by
\[
\begin{aligned}
x_{t+1} &= x_t + \Delta t \cdot v_{x,t} \\
v_{x,t+1} &= v_{x,t} + \Delta t \cdot \left( 1.5 \cdot a_{x,t} - f \cdot v_{x,t} + \epsilon \right), \quad \epsilon \sim \mathcal{N}(0, 0.05^2)
\end{aligned}
\]
where $x_t$ is an $x$ coordinate at timestep $t$, $v_{x,t}$ the velocity at $x$-axis, $f$ the fraction constant which is set to 0 in our setting, and $\epsilon$ is the Gaussian noise with mean zero and standard deviation 0.05. The same rule is applied to the y-axis. The reward is defined as $r = \exp\left(-0.6 \cdot | [x, y] - [0.0, -3.0] |_2 \right)$ where the reward gradually diminishes based on the Euclidean distance from the goal position. If the distance becomes less than 0.25, the episode is considered successful. Each task is defined by the context vector $c = [x_{\text{gate}}, w_{\text{gate}}]$ where $x_{\text{gate}}$ is the center of the door located at $y=0$ and  $w_{\text{gate}}$ the width of the door. On the timestep that the agent passes the $y=0$ and its x-coordinate is out of $\left[x_{\text{gate}} - \frac{w_{\text{gate}}}{2}, x_{\text{gate}} + \frac{w_{\text{gate}}}{2}\right]$, it is considered a collision with the wall. In this case, the agent's position is reset to the collision point, and its velocity is reset to 0.

\subsection{BipedalWalker Environment}
\label{sec:appendix_bipedalwalker}

The BipedalWalker environment is based on the environment BipedalWalkerHardcore provided by OpenAI Gym, as implemented in~\citet{romac2021teachmyagent}. In this environment, the agent is a bipedal robot that walks forward on terrain with evenly spaced stumps of uniform height, using a 4-dimensional continuous action vector to control its left and right hip and knee joints. The agent receives different types of reward and penalty which are (1) forward reward: based on the increase in the agent's horizontal position (2) torque penalty: proportional to the magnitude of the joint torques (3) stability reward: given when the agent maintains a stable hull posture while walking (4) termination penalty: applied when the agent's head touches the ground. The episode ends when the agent reaches the end of the track, exceeds 2000 steps, or the agent's head touches the ground. Observation space is a 24-dimensional continuous vector consisting of 10-dimensional lidar sensor detecting front stumps, 4-dimensional hull state containing information such as angle, angular velocity and horizontal/vertical velocity, 4-dimensional joint angle and 4-dimensional velocity containing left and right hip and knee angle or velocity, and information of whether each leg is contacting the ground (2-dimensional). Each context (task) is defined by the context vector $c=[h_{\text{stump}}, s_{\text{stump}}]$ where $h_{\text{stump}}$ is the height of the stump, $s_{\text{stump}}$ is the interval between two stumps.  

\subsection{Network Structures}
\label{sec:appendix:implementation_details:network_structures}

Our framework employs modular neural networks to approximate each component of the underlying causal structure—state, action, transition, and reward. Each model is instantiated as an ensemble to enable SCM difference approximation through disagreement.

\paragraph{Transition Prediction.}
The transition model approximates the environment dynamics by predicting the next state \(\hat{s}_{t+1}\) from the current state \(s_t\) and action \(a_t\). Each ensemble member is a two-layer MLP: the input \([s_t; a_t]\) is passed through a hidden layer of dimension 32, followed by a ReLU activation and a final linear layer outputting a vector of the same dimension as the state space. The loss is defined as the mean squared error between predicted and actual next states:
\[
\mathcal{L}_{\text{transition}} = \|\hat{s}_{t+1} - s_{t+1}\|_2^2.
\]

\paragraph{Reward Prediction.}
The reward model learns to approximate the reward function \(r(s_t, a_t)\) using a similar architecture to the transition model. It takes \([s_t; a_t]\) as input, processes it through a single hidden layer with ReLU activation, and outputs a scalar reward prediction. The loss is again the mean squared error:
\[
\mathcal{L}_{\text{reward}} = \|\hat{r}_t - r_t\|_2^2.
\]

\paragraph{State and Action Reconstruction.}
To model state and action representations, we employ a \(\beta\)-VAE for each modality. Each encoder maps the input (state or action) to a Gaussian latent distribution parameterized by mean and log-variance vectors. A latent code is sampled via the reparameterization trick and decoded to reconstruct the input. The loss combines reconstruction error and KL divergence:
\[
\mathcal{L}_{\text{VAE}} = \|\hat{x} - x\|_2^2 + \beta \cdot \text{KL}\left(q(z|x) \mid \mathcal{N}(0, I)\right),
\]
where \(x\) is either a state or an action vector. The latent dimensions for state and action VAEs are 32 and 16, respectively. We use a weighting coefficient of $\beta$=4 to balance reconstruction fidelity and latent space regularization.

\paragraph{Ensemble and Uncertainty Estimation.}
All four model types (transition, reward, state VAE, and action VAE) are trained as ensembles of size 10. Disagreement is quantified as the standard deviation of predictions across ensemble members, averaged over a batch of samples. This disagreement serves as a proxy for causal misalignment and is integrated into the task selection criterion.

\paragraph{Optimization.}
Each model in the ensemble is optimized independently using the Adam optimizer with a shared learning rate. During training, the reconstruction or prediction losses are computed and backpropagated per model.
\subsection{Reproducing Baselines}

All baseline methods were integrated from their respective open-source implementations and evaluated under our unified experimental settings. For the PM baselines, we used the implementation available at \url{https://github.com/psclklnk/currot}, and for BW, we adopted the implementation from \url{https://github.com/flowersteam/TeachMyAgent}.
\section{Hyperparameters}
\label{sec:hyperparameters}

\begin{table}[h]
\centering
\caption{\textbf{Hyperparameters used for training each method in each environment.}}
\label{tab:hyperparams}
\resizebox{0.8\textwidth}{!}{
\begin{tabular}{llccc}
\toprule
\textbf{HyperParameter} & & \textbf{Point Mass} & \textbf{BipedalWalker}\\
\midrule
\multicolumn{4}{l}{\textbf{PPO/SAC}} \\
$\gamma$ & & 0.95 & 0.99 \\
$\lambda$ & & 0.99 & 0.95 \\
PPO rollout length & & 4096 & - \\
PPO epochs & & 10 & - \\
PPO minibatches size & & 128 & -  \\
PPO clip range & & 0.2 & - \\
PPO number of workers & & 1 & - \\
PPO max gradient norm & & 0.5 & - \\
PPO value clipping & & False & - \\
PPO Entropy coefficient & & 0.0 & - \\
SAC batch size & & - & 1000 \\
SAC replay buffer size & & - & 2e6 \\
SAC polyak, $\rho$ & & - & 0.995 \\
SAC start steps & & - & 1e4 \\
SAC Entropy coefficient $\alpha$ & & - & 5e-3 \\
Adam learning rate & & 3e-4 & 1e-3 \\
Adam, $\epsilon$ & & 1e-5 & 1e-8 \\
\midrule
\multicolumn{4}{l}{\textbf{CP-DRL}} \\
state weight, $w_{state}$ & & 0 & 100 \\
action weight, $w_{action}$ & & 0 & 20\\
transition probability weight, $w_{transition}$ & & 10 & 100 \\
reward weight, $w_{reward}$ & & 0 & 100\\
Ensemble size, $K$ & & 10 & 10\\
\bottomrule
\end{tabular}
}
\end{table}

\section{Overall Workflow}
\label{sec:appendix_overall_workflow}

This section outlines the overall workflow of CP-DRL. As illustrated in Algorithm~\ref{alg:pseudo_code}, the method iteratively updates a context distribution to guide curriculum learning. At each iteration, the agent samples a batch of tasks, trains its policy, and collects the corresponding trajectories. The disagreement scores for the sampled tasks are computed as defined in Eq.~\ref{eq:disagreement}. This score is combined with episodic return and is used as a cost in CURROT’s optimal transport framework to update the context distribution. This procedure encourages the agent to prioritize causally informative tasks and facilitates efficient learning progression.

\begin{algorithm}
\caption{Causally-Paced Deep Reinforcement Learning (CP-DRL)}
\begin{algorithmic}[1]
\label{alg:pseudo_code}
\STATE \textbf{Input:} Initial context distribution $\hat{p}_{W,0}(\mathbf{c})$, ensemble size $K$, component weights $\{w_{j}\}$, distance bound $\epsilon$, number of curriculum iterations $L$, number of sampled tasks per iteration $M$, episode length $T$
\FOR{$l = 0$ to $L - 1$}
   \STATE \textbf{Agent Improvement:}
   \FOR{$j = 1$ to $M$}
       \STATE Sample context $c_j \sim \hat{p}_{W,l}(\mathbf{c})$
       \STATE Train policy $\pi$ on $c_j$ and collect trajectory $\tau_{c_j} = \{(s_t, a_t, r_t, s_{t+1})\}_{t=1}^{T}$
       \FOR{each causal factor $i \in \{\text{state}, \text{action}, \text{transition}, \text{reward}\}$}
           \STATE Compute disagreement for component $i$ using Eq.~\ref{eq:disagreement}
           \STATE Train an ensemble of $K$ neural networks using loss $\mathcal{L}_i$ (Eq.~\ref{eq:component_losses})
       \ENDFOR
       \STATE Compute episodic return $R_j = \sum_{t=1}^{T} r_{c_j}(s_t, a_t)$
       \STATE Compute causal misalignment score $\text{CM}(c_j)$ using Eq.~\ref{eq:misalignment_score}
   \ENDFOR
   \STATE \textbf{Context Distribution Update:}
   \STATE Update context distribution $\hat{p}_{W,l}(\mathbf{c})$ using CURROT with cost $R_j + \text{CM}(c_j)$
\ENDFOR
\end{algorithmic}
\end{algorithm}

\section{Component-wise Sensitivity Analysis of Disagreement Metrics}
\label{sec:appendix_analysis_disagreement_metrics}

\begin{figure}[h]
    \centering
    \begin{subfigure}{\linewidth}
        \includegraphics[width=\linewidth]{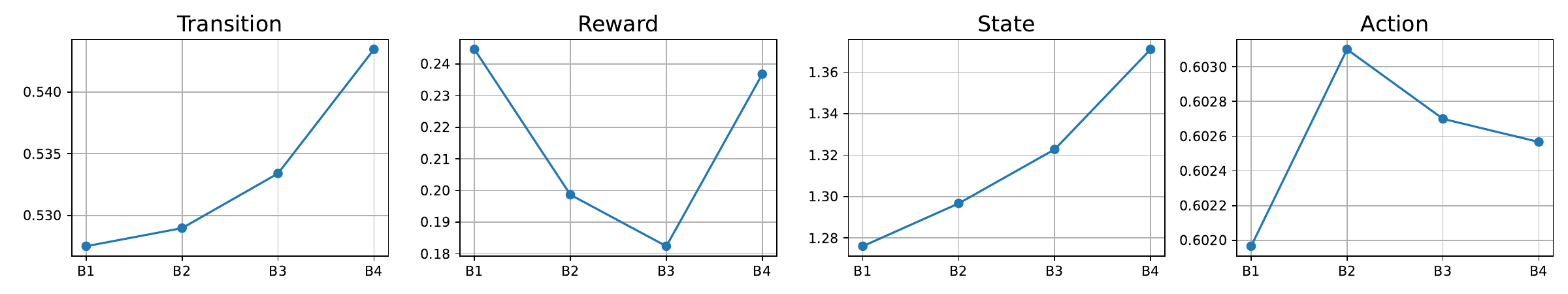}
        \caption{Disagreement trends with increasing block size.}
    \end{subfigure}
    \begin{subfigure}{\linewidth}
        \includegraphics[width=\linewidth]{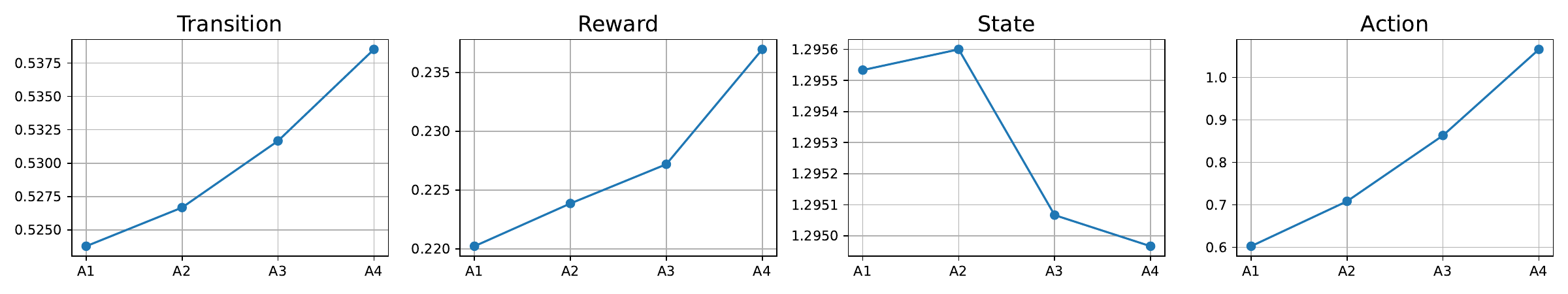}
        \caption{Disagreement trends with increasing action scale.}
    \end{subfigure}
    \begin{subfigure}{\linewidth}
        \includegraphics[width=\linewidth]{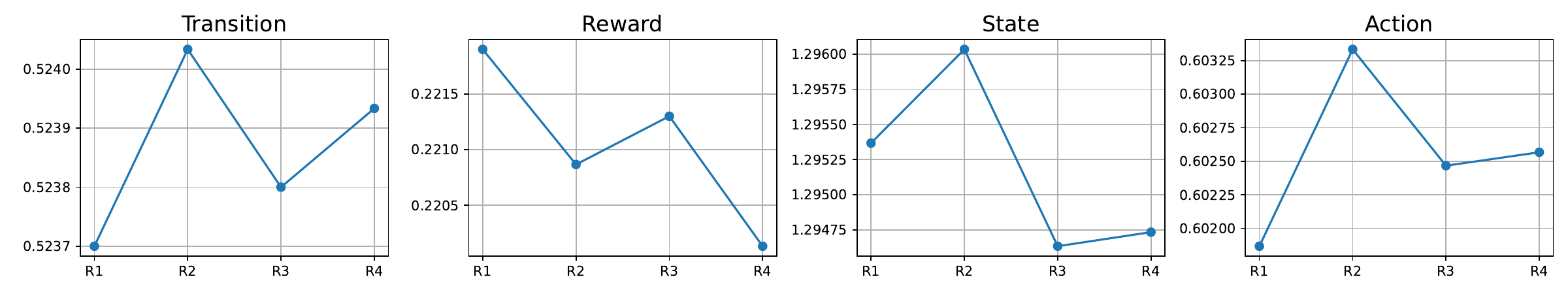}
        \caption{Disagreement trends with increasing reward scale.}
    \end{subfigure}
    \caption{\textbf{Component-wise disagreement under isolated changes to block size, action scale, and reward scale.}}
    \label{fig:cm_additional_analysis}
\end{figure}

To further validate that each disagreement score captures the intended component of causal variation, we conduct a controlled analysis by independently varying block size, action scale, and reward scale, while keeping the other two factors fixed. Specifically, we define four configurations for each factor, resulting in three experimental groups: B1-B4 for block size, A1-A4 for action scale, and R1-R4 for reward scale. All configurations are derived from a base environment corresponding to Task T1, which uses a block size of 0.05, an action scale of 1.0, and a reward weight of 1.0. For each subsequent task (T2 through T4), these parameters are progressively increased, with T4 using a block size of 0.1, an action scale of 4.0, and a reward weight of 4.0.

To isolate the effect of each component, we modify one factor at a time: in the block size group (B1-B4), we vary block size from 0.05 to 0.1 while fixing action scale and reward weight; in the action scale group (A1-A4), we vary action scale while holding block size and reward weight constant; and in the reward scale group (R1-R4), we vary only the reward weight. Figure~\ref{fig:cm_additional_analysis} reports the disagreement scores obtained from each setting.

We observe that transition disagreement increases consistently with both block size and action scale, aligning with the intuition that changes in these factors directly affect the environment's transition dynamics. Similarly, action disagreement steadily grows with increasing action scale, which is expected since the range of agent actions expands. State disagreement is primarily influenced by block size, suggesting that larger blocks alter the spatial configuration of the environment and thereby affect the agent’s state representation. In contrast, reward disagreement exhibits no clear pattern under any of the settings, which we attribute to the sparsity of the reward signal and the use of random trajectories during data collection.

Together, these observations reinforce the interpretation that the proposed modular disagreement metrics serve as meaningful proxies for the underlying causal differences introduced by structural variations in the environment.

\section{When Causal Signals Fail: A Case Study in SGR}
\label{sec:additional_experiments}

\begin{figure}[h]
    \centering
    \includegraphics[width=0.2\linewidth]{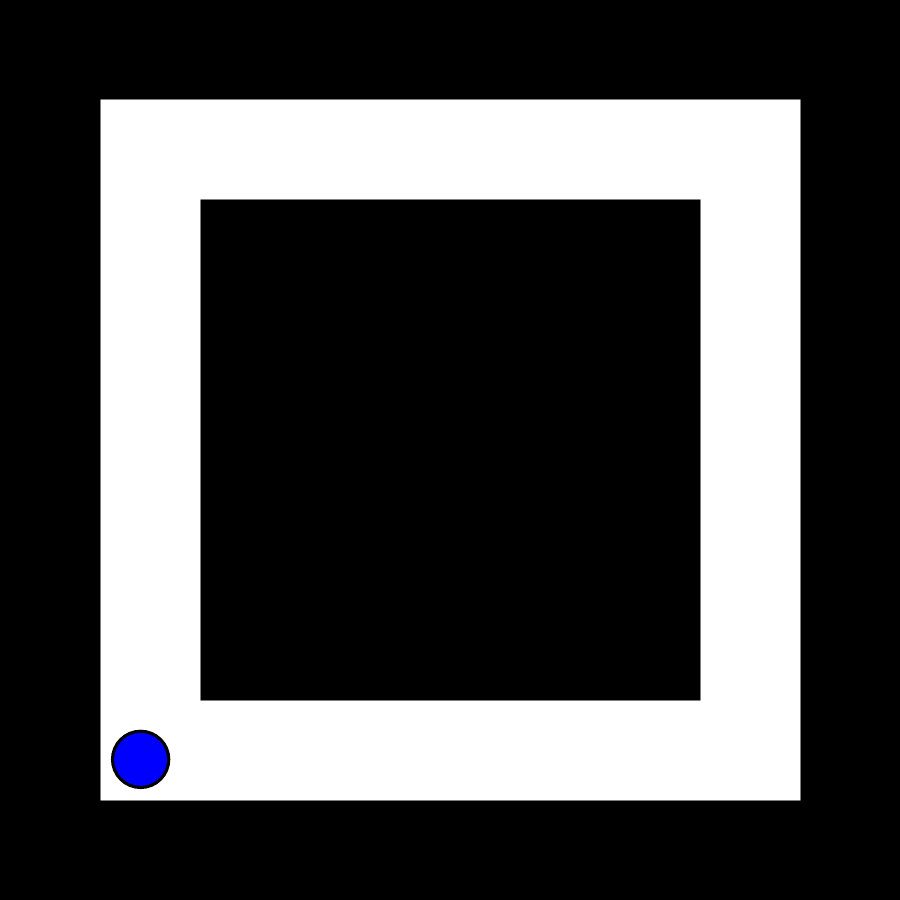}
    \caption{\textbf{Illustration of the Sparse Goal Reaching environment.}}
    \label{fig:enter-label}
\end{figure}

To further investigate the limitations of CP-DRL, we conducted an additional experiment in the Sparse Goal Reaching (SGR) environment. SGR is a 2D continuous maze-like setting adapted from~\citet{florensa2018automatic}, where the agent must navigate to a specified goal location. The environment is bounded within a square region $[-7, 7]^2$, with an impassable square wall occupying the central area $[-5, 5]^2$. The agent's state is represented by a 2D position vector $[x, y]$, and its action is a 2D vector $[a_x, a_y]$ clipped to $[-1, 1]^2$, scaled by 0.3. If the action's norm exceeds 1, it is normalized to unit length before scaling. Collisions with the wall prevent movement, resulting in the agent remaining at its current position.

Each task in this environment is defined by a context vector $c = [x_{\text{goal}}, y_{\text{goal}}, \epsilon] \in \mathbb{R}^3$, where $(x_{\text{goal}}, y_{\text{goal}})$ specifies the goal location and $\epsilon$ is a fixed tolerance threshold, set to $0.05$ in our experiments. The agent receives a reward of $1.0$ upon reaching the goal within the $\epsilon$-radius and a step penalty of $-1.0$ otherwise. Episodes terminate upon successful goal reaching. The agent's initial position is uniformly sampled from $[-7.0, -5.0]^2$, corresponding to the bottom-left region of the maze. To ensure convergence, we trained all baselines for 400 epochs and report results averaged over 10 random seeds.

\begin{figure}[t]
    \centering
    \begin{subfigure}[b]{\linewidth}
        \centering
        \includegraphics[width=1.0\linewidth]{figures/legend_baseline.pdf}
        \vspace{-2.0em}
    \end{subfigure}

    \begin{subfigure}{0.48\linewidth}
        \centering
        \includegraphics[width=\linewidth]{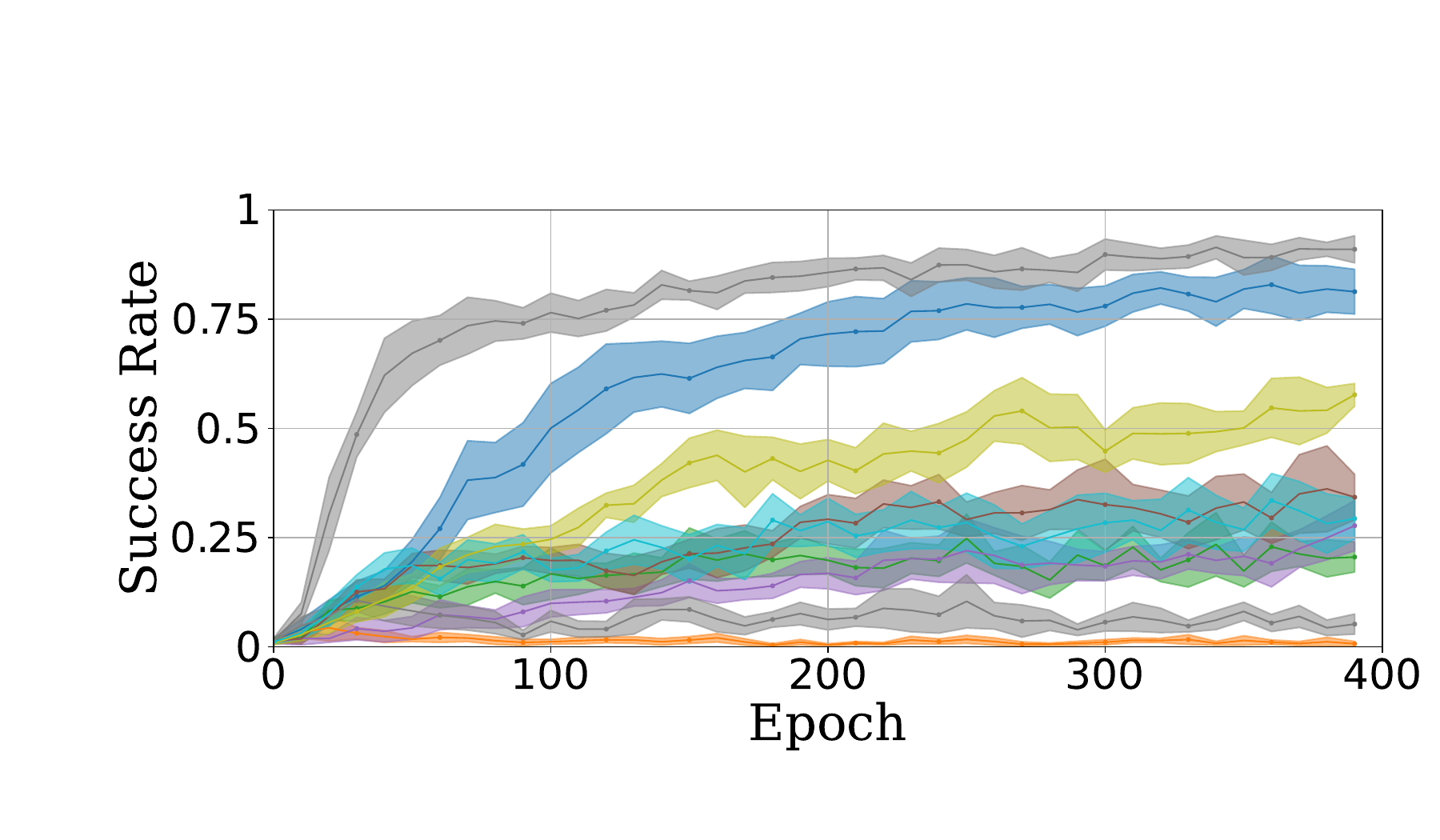}
        \caption{Success Rate}
    \end{subfigure}
    \hfill
    \begin{subfigure}{0.48\linewidth}
        \centering
        \includegraphics[width=\linewidth]{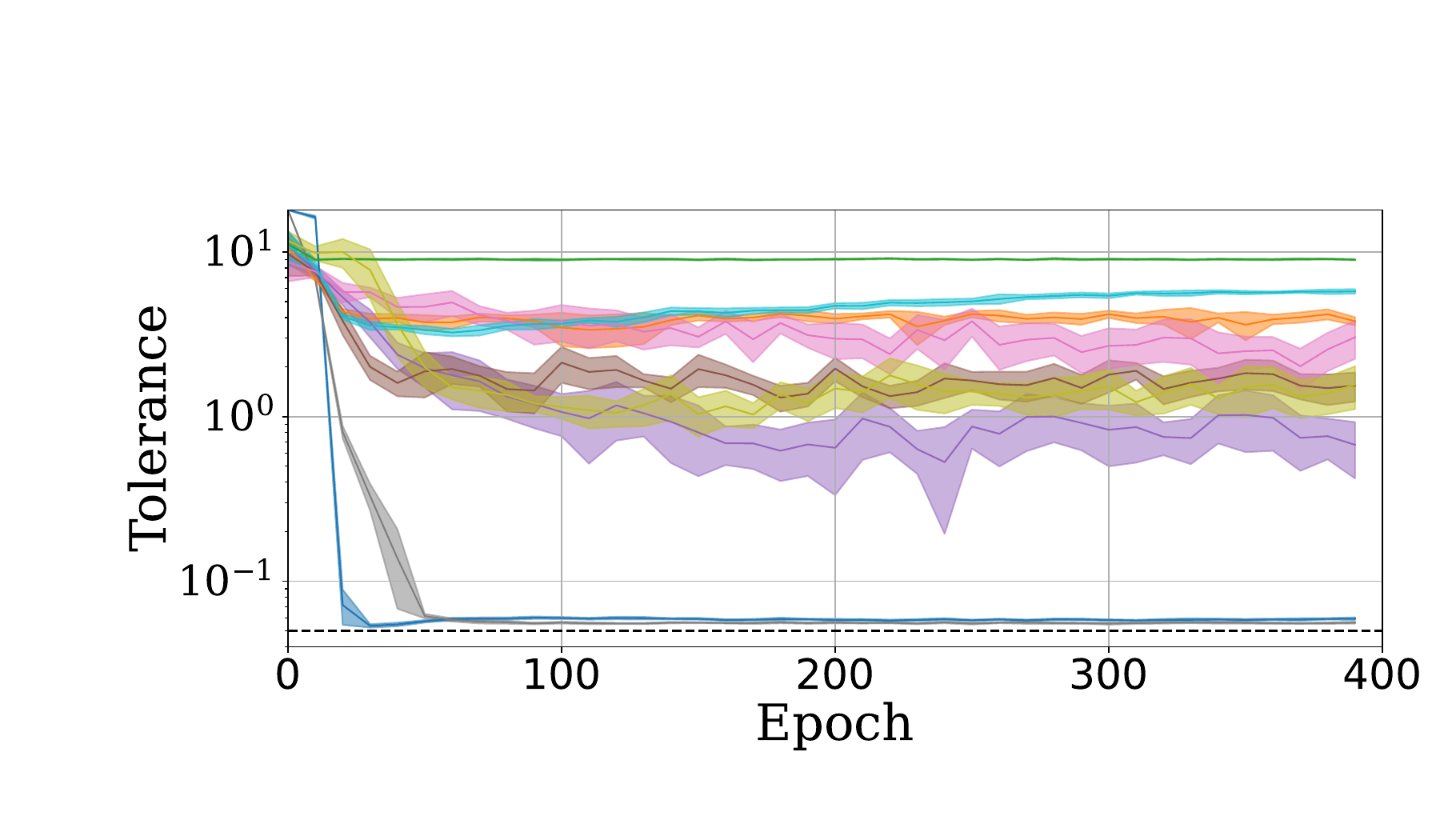}
        \caption{Tolerance}
    \end{subfigure}

    \caption{\textbf{Performance comparison of curriculum methods in Sparse Goal Reaching.} (a) Success rate (b) Median tolerance. All curves show the mean with 95\% confidence intervals over 10 random seeds.}
    \label{fig:sgr_experiment_results}
\end{figure}

As shown in Figure~\ref{fig:sgr_experiment_results}, CP-DRL underperforms compared to CURROT. This performance drop is attributed to the absence of meaningful structural differences between tasks. In the SGR environment, tasks differ only by the goal position, while the underlying state, action, transition dynamics, and reward mechanisms remain invariant. As detailed in Appendix~\ref{sec:appendix_analysis_disagreement_metrics}, when causal structure remains unchanged, the disagreement metrics are primarily driven by noise. Consequently, CP-DRL degenerates into a noisier variant of CURROT. These results highlight a limitation of CP-DRL: when structural variation is minimal or absent, causality-driven curriculum shaping can introduce detrimental stochasticity.
\section{Related Works}
\label{sec:related_works}

Curriculum learning aims to enhance the efficiency of reinforcement learning (RL) and improve performance on the target task by training on a sequence of intermediate tasks guided by various surrogate objectives~\citep{narvekar2020curriculum, klink2022currot}. The learning progress perspective prioritizes tasks that maximize a model's performance gain~\citep{portelas2020teacher, wu2022robust}. The task difficulty perspective focuses on employing tasks of moderate difficulty relative to the current learning state~\citep{florensa2017reverse, florensa2018automatic, huang2022gradient, tzannetos2023proximal, cho2023outcome, sayar2024diffusion}. The regret perspective selects tasks based on their learning potential, quantified by the difference between optimal and current policy rewards~\citep{parker2022evolving, wang2023towards}. The disagreement perspective estimates uncertainty by using multiple Q-functions to measure the variance or mean distance across state-action pairs, and constructs a curriculum that progresses from high to low disagreement to promote exploration and reduce epistemic uncertainty~\citep{zhang2020automatic, cho2023diversify}.

Self-paced curriculum RL~\citep{klink2020sprl} automatically generates and selects tasks based on the agent's learning state without relying on external models. Interpolation-based approaches implement this by gradually shifting the task sampling distribution from an initial to a target distribution~\citep{klink2020scrl, klink2020sprl, klink2021probabilistic, klink2021boosted, klink2022currot, huang2022gradient, tzannetos2023proximal, klink2024benefit}. SPRL~\citep{klink2020sprl} achieves this by using KL Divergence between two task distributions and uses the expected performance constraint. However, it tends to ignore intermediate-level tasks and lacks flexibility in that the task distribution is constrained to be Gaussian or uniform~\citep{klink2022currot, klink2024benefit}. CURROT~\citep{klink2022currot} resolved this problem by viewing the interpolation-based curriculum as an optimal transport problem.

Research efforts that focus on utilizing causal inference in RL consider the causal relationship between variables~\citep{buesing2018woulda, sontakke2021causal, seitzer2021causal, sun2021model, zhu2022invariant}. These works can be divided into two categories, whether it consider structural causal models (SCM), which are Directed Acyclic Graphs that represent causal relationships between variables, as given or not~\citep{zeng2024survey}. \citet{li2024causallyaligned} applied causal inference to curriculum learning, extracting causally aligned states via conditional independence tests and constructing a curriculum by editing them in reverse topological order of actions. Although this approach has demonstrated significant performance gains on specific benchmarks such as Colored Sokoban and Button Maze~\citep{schrader2018gymsokoban, li2024causallyaligned, chevalier2018minigrid}, it still relies on the assumption that the underlying SCM is known in advance, which is often not the case in environments with complex or unknown causal structures~\citep{zeng2024survey}.

Unsupervised Environment Design (UED) approaches utilize an external model to generate suitable environments that help agents acquire the ability to handle diverse situations. It differs from our curriculum learning setting in that it generates the entire environment and focuses on robustness of the policy~\citep{parker2022evolving}. 
\section{Experimental Setup and Reproducibility}
\label{sec:experimental_setup_and_reproducibility}
ALL experiments were performed on a server running Ubuntu 22.04.4 LTS, equipped with an AMD EPYC 7543 32-core CPU, 885GB of RAM, and an 8 NVIDIA A100 GPU. For the BipedalWalker experiments, each run was assigned GPU using Python 3.7.12, PyTorch (v1.13.1) and Gym (v0.17.3).  For the Point Mass experiments, only CPU resources were used with Python 3.8.0, PyTorch (v2.4.1), and Gym (v0.17.3).

\end{document}